\documentclass[letterpaper]{article} 
\usepackage{aaai2027}  
\usepackage[hyphens]{url}  
\usepackage{graphicx} 
\usepackage{natbib}  
\usepackage{caption} 
\usepackage{algorithm}
\usepackage{algorithmic}

\usepackage{newfloat}
\usepackage{listings}
\DeclareCaptionStyle{ruled}{labelfont=normalfont,labelsep=colon,strut=off} 
\floatstyle{ruled}
\newfloat{listing}{tb}{lst}{}
\floatname{listing}{Listing}

\usepackage{booktabs}
\usepackage{amsmath}
\usepackage{diagbox}
\definecolor{bestred}{RGB}{180,35,35}
\definecolor{secondbestblue}{RGB}{35,91,158}
\newcommand{\projectlink}{%
    \leavevmode\pdfstartlink attr{/Border [0 0 0]}
    user{/Subtype /Link /A << /S /URI /URI (https://github.com/RobinY99/MR-IQA-2) >>}%
    \texttt{MR-IQA-2}%
    \pdfendlink
}

\nocopyright 

\title{MR-IQA-2: Faithful Image Quality Reflection via Fine-Grained Credit Assignment}
\author{
    Yuan Li, Youyuan Lin, Chenhui Chu, Shin'ya Nishida
}
\affiliations{
    Graduate School of Informatics, Kyoto University\\
    \raisebox{-0.36\height}{\includegraphics[height=2em]{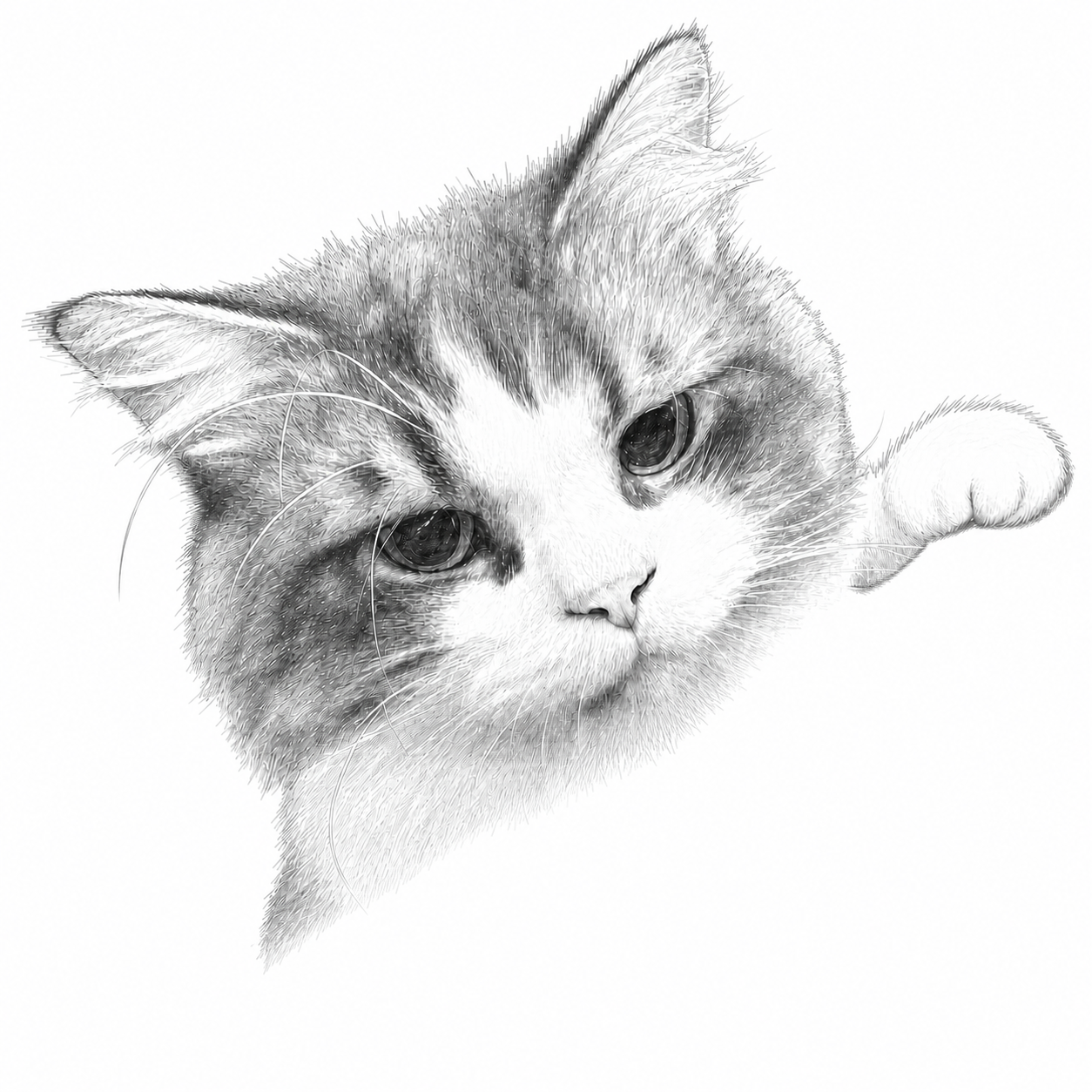}}%
    \hspace{0.3em}\projectlink
}

\begin{document}

\maketitle

\begin{abstract}

Multimodal large language models (MLLMs) have shown great potential to improve image quality assessment (IQA) by making predictions more explainable through increased consistency between quality ratings and their underlying reasoning. However, most existing approaches supervise reasoning only to align it with human-provided quality ratings, paying little attention to whether the reasoning faithfully reflects the image’s actual quality content. Since higher rating accuracy alone does not guarantee faithful reasoning, using a shared reward for both rating and reasoning obscures the source of supervision and may reinforce unfaithful reasoning when a correct rating is obtained by chance.
To improve the faithfulness and reliability of blind IQA, we propose to (1) decouple credit assignment for reasoning and rating, and (2) provide verifiable supervision signals for faithful reasoning.
To this end, we introduce MR-IQA-2, an actor-editor-judge framework for faithful IQA that operationalizes reasoning-editing-reflection. The actor first generates quality reasoning for an input image. Conditioned on this reasoning, the editor revises the image to provide a verifiable visual signal for the identified quality factors. A frozen judge then compares the original and edited images, producing reflective supervision that improves the actor’s quality reasoning.
MR-IQA-2 uses fine-grained credit assignment to decouple reasoning and rating supervision. Judge feedback supervises reasoning, whereas human ratings supervise the predicted rating. Masked token-specific updates distinguish these signals while preserving the causal relation from reasoning to rating.
Across IQA benchmarks, MR-IQA-2 achieves competitive rating alignment with humans. Additionally, visual reflection enables the framework to acquire richer and more faithful visual understanding beyond rating. This faithful visual understanding may inform image-quality optimization and related downstream tasks. Code is available at \projectlink.
\end{abstract}


\section{Introduction}

Image quality assessment (IQA) aims to understand how humans perceive visual quality and to model the relationship between an image and its perceived quality rating.
Prior IQA work has undergone a series of framework transitions as visual content and degradation patterns have become increasingly complex. The goal of IQA is no longer limited to producing a numerical rating. A more important question is whether a model can develop a faithful understanding of image quality. Such understanding should go beyond specific degradations and cover both low-level visual attributes and semantic-level perception.

Early blind image quality assessment (BIQA) methods were largely driven by degradation-centered assumptions. In this setting, image quality was often associated with the strength of synthetic distortions, and the learning objective was commonly formulated as degradation-scale estimation. Methods such as ARNIQA~\citep{arniqa}, TOPIQ~\citep{topiq}, and LIQE~\citep{liqe} represent important attempts to learn quality-aware representations under this paradigm.

With the emergence of practical IQA benchmarks such as KonIQ-10k~\citep{koniq} and SPAQ~\citep{spaq}, the focus of BIQA has gradually shifted. Real-world images contain diverse content, complex capture conditions, aesthetic factors, and semantic preferences. These factors cannot be fully described by synthetic distortion types. As a result, BIQA has moved from estimating degradation intensity toward understanding the image itself.

Recent multimodal approaches further extend this trend. Works such as Q-Instruct~\citep{q-instruct}, Q-Insight~\citep{q-insight}, and VisualQuality-R1~\citep{Visualquality-r1} introduce language, criteria, and human-like reasoning into BIQA. These methods allow models to evaluate images through quality dimensions, ranking preferences, and explanatory outputs. This transition improves interpretability and generalization via criteria closer to human perception. But it also raises a deeper question: does the generated reasoning faithfully reflect the actual factors that determine image quality?

This question is critical for BIQA based on multimodal large language models (MLLMs). MLLMs can produce fluent and plausible quality explanations, but plausible language does not guarantee faithful visual reasoning. A model may attribute low quality to reasonable-sounding factors while failing to identify the true visual limitations. Causal reasoning collapse, unverifiable explanations, and deviations from human perception can therefore appear in quality assessment. Previous studies, including BRIQA~\citep{briqa} and H-IQA~\citep{h-iqa}, have attempted to analyze reasoning failures, improve reasoning consistency, and align IQA with human perception. However, the reasoning process itself remains difficult to verify.

Recent methods, including Zoom-IQA~\citep{zoom-iqa} and
Tool-IQA~\citep{tool-iqa}, move BIQA toward tool-augmented reasoning through
region-aware inspection. These methods show that
additional visual observations help models inspect image details more reliably.
Nevertheless, they focus on observation and grounding rather than verifying
whether the proposed factors actually limit image quality. It therefore remains
unclear whether a model has identified the factors that genuinely constrain
overall image quality.

Another persistent issue in reinforcement learning (RL)-based BIQA methods,
including Q-Insight~\citep{q-insight},
VisualQuality-R1~\citep{Visualquality-r1},
MR-IQA~\citep{mr-iqa},
Zoom-IQA~\citep{zoom-iqa}, and
Tool-IQA~\citep{tool-iqa}, is that reasoning supervision is often determined
by rating performance. Consequently, incorrect reasoning may still receive
high rewards when the rating is accurate, potentially undermining reasoning
faithfulness.

In this work, we decompose the learning of faithful quality reasoning into two
tasks: (1) decoupling reasoning and rating supervision and (2) constructing an
independent supervision signal for reasoning. For the first task, we propose
fine-grained credit assignment. We use masked token-specific updates so that
better reasoning or a more accurate rating is rewarded accordingly within its
corresponding output while preserving causality. The second task is therefore to construct a
reliable supervision signal for reasoning. We use visual reflection to estimate
this signal from observed quality changes. A reasoning proposal receives higher
credit when its corresponding edit alleviates the identified quality-limiting
factors. Finally, we integrate these two mechanisms into an actor-editor-judge
framework termed MR-IQA-2.

Prior work primarily pursues stronger rating performance, as exemplified by MR-IQA~\citep{mr-iqa}, or richer reasoning processes, as in Zoom-IQA~\citep{zoom-iqa} and Tool-IQA~\citep{tool-iqa}. In contrast, MR-IQA-2 investigates whether the generated reasoning faithfully identifies quality-limiting factors and yields visual understanding that can inform broader downstream tasks, \mbox{including image editing and visual recognition}.

\begin{figure*}[!t]
    \centering
    \includegraphics[
        width=\textwidth,
        height=0.68\textheight,
        keepaspectratio,
        pagebox=cropbox
    ]{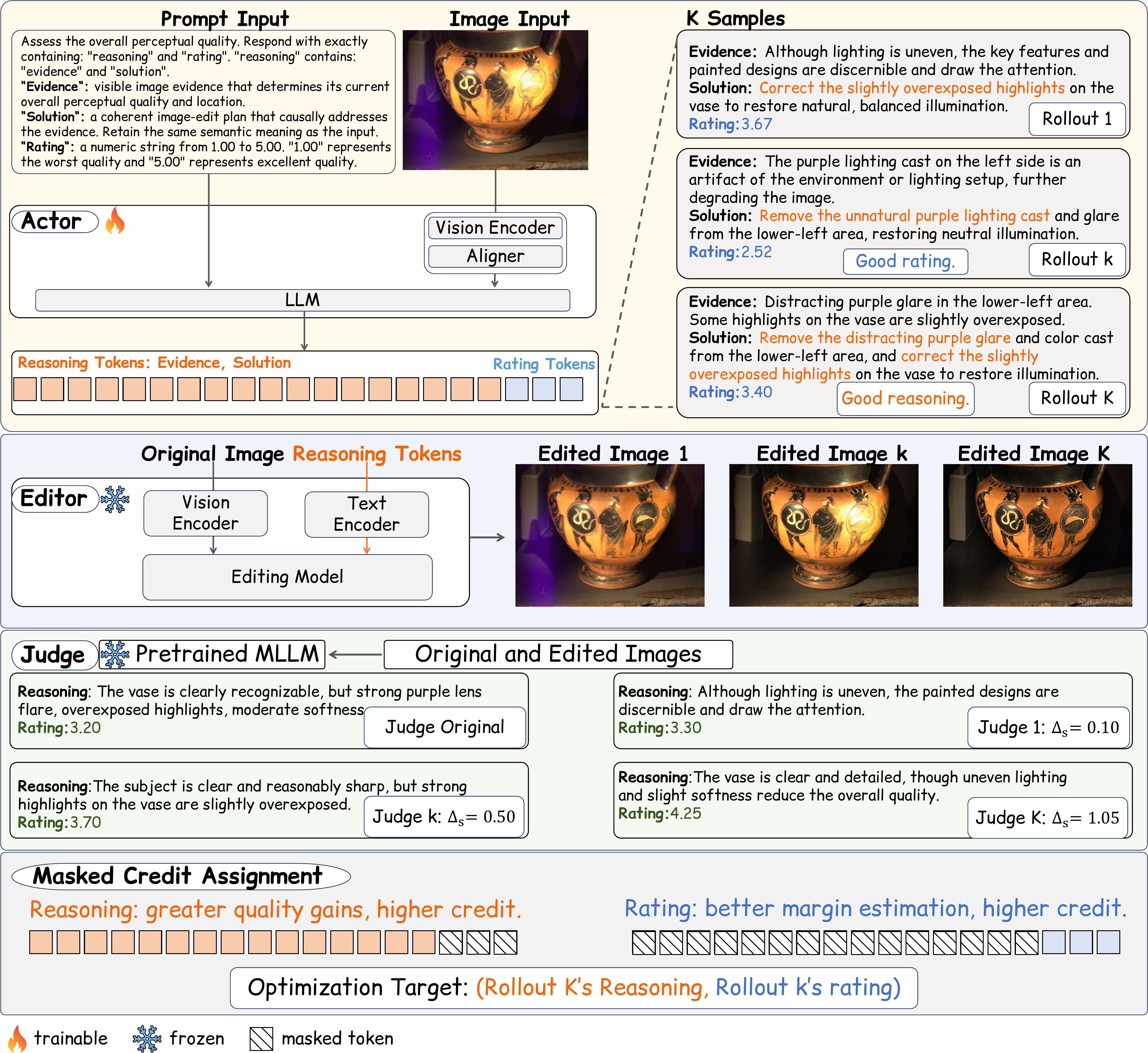}
    \caption{Overview of MR-IQA-2. Prior MLLM-based IQA produces an
    explanation and a rating without visually verifying the explanation.
    Our actor converts a quality hypothesis into multiple targeted
    low-level edits. The independent judge compares the resulting
    quality changes and returns evidence-based feedback. Only visually
    supported interventions reinforce the actor, yielding faithful
    reasoning and a deeper understanding of causal quality-limiting
    factors. Here, $\Delta s$ denotes the Judge-score change between an edited
    image and the original image, as defined in
    Eq.~\eqref{eq:judge-quality-change}.}
    \label{fig:overview}
\end{figure*}

The main contributions are summarized as follows:
\begin{itemize}
    \item We introduce fine-grained credit assignment for the reasoning--rating structure in BIQA. Masked token-specific updates decouple reasoning and rating supervision, assigning signal to its corresponding output while preserving the causal relation from reasoning to rating.

    \item We construct a verifiable supervision signal for faithful quality reasoning through visual reflection. Reasoning-conditioned editing serves as a visual intervention, while a frozen judge evaluates the observed quality change and assigns credit to the corresponding reasoning.

    \item We integrate these two mechanisms into MR-IQA-2, a modular actor-editor-judge framework that operationalizes reasoning-editing-reflection. The actor, editor, and judge can be independently instantiated or replaced for different models and downstream settings.
\end{itemize}

\section{Related Work}

\subsection{Faithful Quality Reasoning with MLLMs.}

The adoption of MLLMs has improved the interpretability of BIQA by enabling language-based image understanding and quality explanations. However, faithful quality reasoning remains a key challenge. Early methods such as DepictQA~\citep{depictqa} and Q-Instruct~\citep{q-instruct} mainly rely on supervised fine-tuning (SFT), which encourages plausible explanation patterns. Constrained by their training data, these models may overfit templated responses and exhibit limited logical coherence. Later RL-based frameworks, including Q-Insight~\citep{q-insight}, VisualQuality-R1~\citep{Visualquality-r1}, H-IQA~\citep{h-iqa}, and MR-IQA~\citep{mr-iqa}, optimize generated reasoning using rating supervision. By strengthening the model's underlying perceptual capabilities, these methods mitigate template overfitting, produce richer and more consistent reasoning, and improve rating alignment. Nevertheless, the reasoning itself is generated by the MLLM backbone without a reliable reasoning-specific supervision signal, making its faithfulness difficult to verify. Our framework instead evaluates reasoning faithfulness through visual intervention by verifying whether edits derived from the identified reasoning factors lead to observable improvements in image quality.

%

\subsection{Tool-Augmented IQA Methods}

Before MLLM-based BIQA, degradation modeling and task-specific image processing were already used to derive quality cues. ARNIQA~\citep{arniqa} applies predefined degradation operators to construct contrastive views and learn distortion-aware representations. More recently, Zoom-IQA~\citep{zoom-iqa} introduces region-aware zooming for detail inspection. Tool-IQA~\citep{tool-iqa} further employs a magnifier and a gamma corrector to provide additional visual evidence for quality rating. Although these tools enable richer and more localized inspection, the resulting reasoning is still evaluated mainly through rating performance. Rather than enriching the visual evidence during reasoning for rating optimization, our design provides direct supervision for the faithfulness of quality reasoning.

\section{Methods}
\label{sec:methods}

\subsection{MR-IQA-2 Framework Overview}
In this work, we propose MR-IQA-2 (Figure~\ref{fig:overview}), an actor-editor-judge framework for faithful IQA. MR-IQA-2 makes quality reasoning verifiable rather than assessing it solely through rating performance. The actor is a general MLLM-based BIQA model that produces interpretable quality reasoning and a rating. A frozen editor, initialized from an image-editing diffusion model, applies edits conditioned on the reasoning. A frozen MLLM judge then evaluates the quality change between the original and edited images. The observed change provides a reasoning-specific supervision signal for the actor. The following sections introduce (1) the interaction among the actor, editor, and judge, (2) fine-grained credit assignment, and (3) the training optimization procedure.

\subsection{Actor-Editor-Judge Trajectory}
\label{sec:actor-editor-judge-trajectory}
This section describes the interaction flow among the Actor, Editor, and Judge.
For a local batch of $N$ images, $i\in\{1,\ldots,N\}$ indexes an image, $K$
denotes the number of Actor samples per image, and $k\in\{1,\ldots,K\}$ indexes
one sample.
Within each training trajectory, superscript $b\in\{0,1\}$ denotes the
original-image and post-editing stages, respectively. \\
\textbf{Actor} receives an original image $I_i^0$ and uses a prompt $P_A$ to
generate quality reasoning and a rating:
\begin{equation}
    a_{i,k}^0
    =
    \left(r_{i,k}^0,q_{i,k}^0\right)
    \sim
    \pi_{\mathrm{old}}\!\left(\cdot\mid I_i^0,P_A\right),
    \label{eq:actor-initial-rollout}
\end{equation}
where $\pi_{\mathrm{old}}$ denotes the original Actor policy used to generate
the rollouts, and $r_{i,k}^0$ and $q_{i,k}^0$ denote reasoning and rating. \\
\textbf{Editor} corrects the quality factors identified by the Actor:
\begin{equation}
    I_{i,k}^1
    =
    E_{\phi_E}\!\left(I_i^0,r_{i,k}^0;P_E\right),
    \qquad \phi_E\ \text{fixed}.
    \label{eq:editor-intervention}
\end{equation}
where $P_E$ is a fixed editing prompt template and $I_{i,k}^1$ is the edited
output.\\
\textbf{Judge} independently rates the original and edited images:
\begin{equation}
    s_i^0
        =J_{\phi_J}\!\left(I_i^0\right),
    \qquad
    s_{i,k}^1
        =J_{\phi_J}\!\left(I_{i,k}^1\right).
    \label{eq:judge-evaluation}
\end{equation}
The resulting trajectory is
\begin{equation}
    \mathcal{T}_{i,k}
    =
    \left(
    I_i^0,a_{i,k}^0,I_{i,k}^1,s_i^0,s_{i,k}^1
    \right).
    \label{eq:actor-editor-judge-trajectory}
\end{equation}
The Judge-observed quality change is
\begin{equation}
    \Delta s_{i,k}=s_{i,k}^1-s_i^0.
    \label{eq:judge-quality-change}
\end{equation}

\subsection{Fine-Grained Credit Assignment}
\paragraph{Motivation.}
As discussed in the Introduction, higher rating accuracy alone does not
guarantee faithful reasoning. Moreover, assigning a shared reward to reasoning
and rating obscures the source of supervision. We therefore define separate
rewards for reasoning, rating, and format, and use each reward to update its
corresponding output tokens.

\paragraph{Reasoning reward.}
For each image, the Actor samples $K$ reasoning--rating outputs indexed by
$k$. Each reasoning proposal is evaluated with the frozen Editor and
Judge. The editing prompt restricts interventions to low-level quality
attributes while preserving image content. We measure reasoning faithfulness
by the resulting Judge-score improvement:
\begin{equation}
    R_{i,k}^{\mathrm{reasoning}}
    =
    \Delta s_{i,k}.
    \label{eq:reasoning-reward}
\end{equation}
Under these controlled conditions, reasoning that produces a larger quality
improvement receives a higher reward.

\paragraph{Rating reward.}
Following MR-IQA~\citep{mr-iqa}, we supervise rating through relative quality
margins. Let $y_i$ be the human mean opinion score (MOS). For another image
$j\ne i$ in the same batch, the scale-controlled margin error is
\begin{equation}
    z_{i,k,j}^{\mathrm{rating}}
    =
    \frac{
    \left(q_{i,k}^0-\frac{1}{K}\sum_{k'=1}^{K}q_{j,k'}^0\right)
    -
    \left(y_i-y_j\right)
    }{\tau_{ij}},
    \label{eq:rating-margin-error}
\end{equation}
where $\tau_{ij}>0$ controls the margin-error scale. Using the L2 estimator, we
average the $N-1$ pairwise rewards:
\begin{equation}
    R_{i,k}^{\mathrm{rating}}
    =
    \frac{1}{N-1}
    \sum_{\substack{j=1\\j\ne i}}^{N}
    \mathrm{e}^{-\frac{1}{2}\left(z_{i,k,j}^{\mathrm{rating}}\right)^2}.
    \label{eq:rating-reward}
\end{equation}

\paragraph{Format reward.}
The Actor output must be a valid JSON object containing the ordered fields
\textbf{reasoning} and \textbf{rating}. Let $\mathcal{F}$ denote this output
contract. The format reward is
\begin{equation}
    R_{i,k}^{\mathrm{format}}
    =
    \mathbf{1}\!\left[a_{i,k}^0\in\mathcal{F}\right].
    \label{eq:format-reward}
\end{equation}

\paragraph{Masked credit assignment.}
Let $c$ index a supervision type. For token $t$, $\chi_{i,k,t}^c$ is its binary
mask, and $\Omega_{i,k}^c$ is the selected token set:
\begin{equation}
\begin{aligned}
    \mathcal{C}
        &=
        \{\mathrm{reasoning},\mathrm{rating},\mathrm{format}\},\\
    \Omega_{i,k}^c
        &=
        \left\{t\,\middle|\,\chi_{i,k,t}^c=1\right\},
        \qquad c\in\mathcal{C}.
\end{aligned}
    \label{eq:credit-token-set}
\end{equation}
Each reward is normalized independently within the $K$ samples of image $i$:
\begin{equation}
    A_{i,k}^c
        =\frac{R_{i,k}^c-\mu_i^c}{\sigma_i^c+\epsilon_A}.
    \label{eq:reward-wise-advantage}
\end{equation}
Here, $\mu_i^c$ and $\sigma_i^c$ denote the corresponding group mean and sample
standard deviation. The masked advantage for each supervision type is
\begin{equation}
    \widetilde A_{i,k,t}^c
    =
    \chi_{i,k,t}^c A_{i,k}^c.
    \label{eq:token-level-credit}
\end{equation}
Thus, reasoning and rating rewards update only their corresponding fields,
whereas the format reward applies to the complete output.

\subsection{Masked Credit with GRPO}
We optimize the Actor with Group Relative Policy Optimization
(GRPO)~\citep{shao2024deepseekmath}. Let $\mathcal{I}_c$ contain the valid
sampled outputs for supervision type $c$. Using the masked advantage in
Eq.~\eqref{eq:token-level-credit}, the masked GRPO loss is
\begin{equation}
\begin{aligned}
    \mathcal{L}_{\mathrm{GRPO}}(\theta)
        &=-\sum_{c\in\mathcal{C}}\frac{1}{|\mathcal{I}_c|}
        \sum_{(i,k)\in\mathcal{I}_c}
        \frac{1}{|\Omega_{i,k}^c|}
        \sum_{t\in\Omega_{i,k}^c}\omega_{i,k,t}\\
        &\quad{}\times
        \min\left(
            \rho_{i,k,t}\widetilde A_{i,k,t}^c,
            \rho_{i,k,t}^{\mathrm{clip}}\widetilde A_{i,k,t}^c
        \right).
\end{aligned}
    \label{eq:masked-grpo-loss}
\end{equation}
Here, $\rho_{i,k,t}$ is the likelihood ratio between the current Actor
$\pi_\theta$ and the rollout policy $\pi_{\mathrm{old}}$.
$\rho_{i,k,t}^{\mathrm{clip}}$ clips this ratio to
$[1-\epsilon_{\mathrm{low}},1+\epsilon_{\mathrm{high}}]$.
$\omega_{i,k,t}$ corrects the token-level mismatch between the rollout
distribution and $\pi_{\mathrm{old}}$ and is capped by $\omega_{\max}$.

\begin{table*}[!t]
\centering
{\small
\setlength{\tabcolsep}{2pt}
\begin{tabular}{@{}l cc cc cc cc cc cc cc@{}}
\toprule
& \multicolumn{6}{c}{Authentic} & \multicolumn{2}{c}{AI-generated} &
\multicolumn{4}{c}{Synthetic} & \multicolumn{2}{c}{Average} \\
\cmidrule(lr){2-7} \cmidrule(lr){8-9} \cmidrule(lr){10-13}
\cmidrule(lr){14-15}
& \multicolumn{2}{c}{KonIQ} & \multicolumn{2}{c}{SPAQ} &
\multicolumn{2}{c}{LIVE-W} & \multicolumn{2}{c}{AGIQA-3K} &
\multicolumn{2}{c}{KADID-10k} & \multicolumn{2}{c}{CSIQ} &
\multicolumn{2}{c}{Avg.} \\
\cmidrule(lr){2-3} \cmidrule(lr){4-5} \cmidrule(lr){6-7}
\cmidrule(lr){8-9} \cmidrule(lr){10-11} \cmidrule(lr){12-13}
\cmidrule(lr){14-15}
Method & PLCC & SRCC & PLCC & SRCC & PLCC & SRCC & PLCC & SRCC &
PLCC & SRCC & PLCC & SRCC & PLCC & SRCC \\
\midrule
\multicolumn{15}{@{}l}{\textit{Hand-crafted}} \\
NIQE~\citeyearpar{NIQE}
& 0.533 & 0.530 & 0.679 & 0.664 & 0.493 & 0.449 & 0.560 & 0.533
& 0.468 & 0.405 & 0.718 & 0.628 & 0.575 & 0.535 \\
BRISQUE~\citeyearpar{BRISQUE}
& 0.225 & 0.226 & 0.490 & 0.406 & 0.361 & 0.313 & 0.541 & 0.497
& 0.429 & 0.356 & 0.740 & 0.556 & 0.464 & 0.392 \\
\midrule
\multicolumn{15}{@{}l}{\textit{Deep-learning-based}} \\
NIMA~\citeyearpar{nima}
& 0.896 & 0.859 & 0.838 & 0.856 & 0.814 & 0.771 & 0.715 & 0.654
& 0.532 & 0.535 & 0.695 & 0.649 & 0.748 & 0.721 \\
DBCNN~\citeyearpar{dbcnn}
& 0.884 & 0.875 & 0.812 & 0.806 & 0.773 & 0.730 & 0.641 & 0.648
& 0.497 & 0.484 & 0.586 & 0.572 & 0.699 & 0.686 \\
MUSIQ~\citeyearpar{musiq}
& 0.924 & 0.929 & 0.868 & 0.863 & 0.789 & 0.830 & 0.722 & 0.630
& 0.575 & 0.556 & 0.771 & 0.710 & 0.775 & 0.753 \\
MANIQA~\citeyearpar{yang2022maniqa}
& 0.849 & 0.834 & 0.768 & 0.758 & 0.849 & 0.832 & 0.723 & 0.636
& 0.499 & 0.465 & 0.623 & 0.627 & 0.719 & 0.692 \\
CLIP-IQA+~\citeyearpar{wang2023exploring}
& 0.909 & 0.895 & 0.866 & 0.864 & 0.832 & 0.805 & 0.736 & 0.685
& 0.653 & 0.654 & 0.772 & 0.719 & 0.795 & 0.770 \\
\midrule
\multicolumn{15}{@{}l}{\textit{MLLM-based: SFT training}} \\
C2Score~\citeyearpar{zhu2024adaptive}
& 0.923 & 0.910 & 0.867 & 0.860 & 0.786 & 0.772 & 0.777 & 0.671
& 0.500 & 0.453 & 0.735 & 0.705 & 0.765 & 0.729 \\
Q-Align~\citeyearpar{q-align}
& 0.941 & \textcolor{secondbestblue}{0.940}
& 0.886 & 0.887 & 0.853 & 0.860 & 0.772 & 0.735
& 0.674 & 0.684 & 0.671 & 0.737 & 0.800 & 0.807 \\
DeQA~\citeyearpar{deqa}
& \textcolor{bestred}{0.953} & \textcolor{bestred}{0.941}
& 0.895 & 0.896 & 0.892 & \textcolor{secondbestblue}{0.879}
& 0.809 & 0.729
& 0.694 & 0.687 & 0.787 & 0.744
& \textcolor{secondbestblue}{0.838} & 0.813 \\
\midrule
\multicolumn{15}{@{}l}{\textit{MLLM-based: RL training}} \\
Q-Insight~\citeyearpar{q-insight}
& 0.918 & 0.895 & \textcolor{bestred}{0.903}
& \textcolor{bestred}{0.903} & 0.870 & 0.839
& \textcolor{secondbestblue}{0.816} & \textcolor{secondbestblue}{0.766}
& \textcolor{bestred}{0.702} & \textcolor{bestred}{0.702}
& 0.685 & 0.640 & 0.816 & 0.791 \\
VQ-R1~\citeyearpar{Visualquality-r1}
& 0.886 & 0.919 & 0.867 & 0.887 & 0.817 & 0.869 & 0.744 & 0.718
& 0.635 & 0.640 & 0.709 & 0.721 & 0.776 & 0.792 \\
MR-IQA~\citeyearpar{mr-iqa}
& \textcolor{secondbestblue}{0.949} & 0.931
& 0.892 & 0.897 & \textcolor{bestred}{0.899}
& \textcolor{bestred}{0.883} & 0.804 & 0.732 & 0.672 & 0.683
& 0.767 & 0.732 & 0.831 & 0.810 \\
\midrule
\multicolumn{15}{@{}l}{\textit{MLLM-based: Tool-augmented training}} \\
Zoom-IQA$^{\dagger}$~\citeyearpar{zoom-iqa}
& 0.938 & 0.922
& \textcolor{secondbestblue}{0.902} & \textcolor{secondbestblue}{0.900}
& 0.887 & 0.870
& \textcolor{secondbestblue}{0.816} & 0.765
& \textcolor{secondbestblue}{0.701} & \textcolor{secondbestblue}{0.700}
& 0.797 & 0.754
& \textcolor{bestred}{0.840} & \textcolor{bestred}{0.819} \\
\textbf{MR-IQA-2 (Ours)}
& 0.937 & 0.917
& 0.900 & 0.899
& \textcolor{secondbestblue}{0.893} & 0.863
& 0.809 & 0.739
& 0.667 & 0.669
& \textcolor{secondbestblue}{0.824} & \textcolor{secondbestblue}{0.785}
& \textcolor{secondbestblue}{0.838} & 0.812 \\
\textbf{MR-IQA-2$^{*}$ (Ours)}
& 0.925 & 0.904
& 0.901 & \textcolor{secondbestblue}{0.900}
& 0.881 & 0.847
& \textcolor{bestred}{0.826} & \textcolor{bestred}{0.768}
& 0.665 & 0.676
& \textcolor{bestred}{0.840} & \textcolor{bestred}{0.801}
& \textcolor{bestred}{0.840} & \textcolor{secondbestblue}{0.816} \\
\bottomrule
\end{tabular}
}
\caption{\textbf{Rating performance comparison.} Each dataset reports
PLCC$\uparrow$ and SRCC$\uparrow$ between predicted and ground-truth ratings.
\textcolor{bestred}{Red} and \textcolor{secondbestblue}{blue} denote the best
and second-best results, respectively. Baseline entries use reported results, except that VQ-R1 is reproduced with
Qwen3-VL-2B~\citep{qwen3vl} because its original results were obtained under a
different training protocol. Both MR-IQA-2 variants use Qwen3.5-4B;
the unstarred row uses the final E5 credit-mask checkpoint, while
MR-IQA-2$^{*}$ freezes the vision encoder and aligner during training.
Zoom-IQA$^{\dagger}$ uses additional annotations and combines
SFT with RL; all other trainable methods use the same KonIQ training split.}
\label{tab:main}

{\small
\setlength{\tabcolsep}{6pt}
\begin{tabular*}{\textwidth}{@{\extracolsep{\fill}}l|rr|rr|c@{}}
\toprule
& \multicolumn{2}{c|}{\textbf{Pretrained Judge}}
& \multicolumn{2}{c|}{\textbf{Q-Insight as Judge}}
& \multicolumn{1}{c}{\textbf{GPT-5.6 Sol as Judge (\%)}} \\
\midrule
\diagbox[width=10.5em,height=2.8\baselineskip]{\textbf{Actor}}{\textbf{Editor}}
& \multicolumn{1}{c}{\textbf{FLUX.2 [klein]}}
& \multicolumn{1}{c|}{\textbf{Mage-Flow}}
& \multicolumn{1}{c}{\textbf{FLUX.2 [klein]}}
& \multicolumn{1}{c|}{\textbf{Mage-Flow}}
& \multicolumn{1}{c}{\textbf{FLUX.2 [klein]}} \\
\midrule
Qwen3.5-4B (Baseline) & $-0.137$ & $-0.144$ & $-0.130$ & $-0.080$ & $4.19\%$ \\
Q-Insight & $+0.213$ & $+0.142$ & $+0.138$ & $+0.139$ & $14.41\%$ \\
\textbf{MR-IQA-2 (Ours)} & $\mathbf{+0.789}$ & $\mathbf{+0.468}$
& $\mathbf{+0.561}$ & $\mathbf{+0.376}$ & $\mathbf{81.40\%}$ \\
\bottomrule
\end{tabular*}
}
\caption{\textbf{Reasoning performance validation.} The first four columns
report mean quality gain $\Delta s$ (higher is better;
Eq.~\eqref{eq:judge-quality-change}); GPT-5.6 Sol uses 3-Alternative Forced Choice (3AFC). We compare
the Qwen3.5-4B baseline~\citep{qwen35}, Q-Insight~\citep{q-insight}, and
MR-IQA-2 (ours). The 4B four-step LoRA Editors are FLUX.2
[klein]~\citep{flux2} and Mage-Flow~\citep{zhang2026mageflow}. Judges are our
pretrained Qwen3.5-4B Judge~\citep{qwen35}, Q-Insight~\citep{q-insight}, and
GPT-5.6 Sol~\citep{openai2026gpt56sol}. Evaluation uses a fixed
5,866-image subset formed by randomly selecting 1,000 images from each test set,
except CSIQ, for which all 866 images are used.}
\label{tab:cross-tool-reasoning}
\end{table*}

\section{Experiments}
\label{sec:experiments}

\subsection{Experimental Settings}
\label{sec:experimental-settings}

\paragraph{Datasets.}
We train all controlled models on the training subset of
the KonIQ-10k split~\citep{koniq}, which contains $7{,}046$
in-the-wild images at $512{\times}384$ resolution. We evaluate on the
KonIQ-10k test split (2,010 images) as the in-distribution authentic benchmark.
For out-of-distribution (OOD) evaluation, we use the authentic distortion
datasets SPAQ (11,125 images)~\citep{spaq} and LIVE In the Wild
(LIVE-W; 1,162 images)~\citep{live-w}, as well as the AI-generated image
quality dataset AGIQA-3K (2,982 images)~\citep{agiqa}. We further include the
synthetic distortion datasets KADID-10k (10,125 images)~\citep{kadid} and CSIQ
(866 images)~\citep{csiq}.

\paragraph{Model Backbones.}
The \textbf{Actor} is initialized from Qwen3.5-4B~\citep{qwen35}, selected for
its balance of visual reasoning and efficiency. The \textbf{Editor} uses
FLUX.2 [klein] 4B~\citep{flux2} with a four-step LoRA for fast and efficient
editing. It runs in BF16 with classifier-free guidance $1.0$ and sigma schedule
$[1.0,0.75,0.5,0.25]$. The \textbf{Judge} uses a Qwen3.5-4B checkpoint trained
for 5 epochs on KonIQ-7k with rating supervision, selected for the same balance
of visual reasoning and efficiency. On a single NVIDIA RTX A6000, the Editor's
mean inference time is $0.902$\,s per image, while Judge inference takes $0.800$\,s per image.
During GRPO optimization, only the Actor is updated; the frozen Editor and
Judge provide offline supervision and receive no gradients.

\paragraph{Implementation Details.}
We train Qwen3.5-4B~\citep{qwen35} for five epochs on eight 48-GB NVIDIA RTX
A6000 GPUs. Each rank processes $N=6$ images and samples $K=6$ completions per
image. The maximum completion length is 192 tokens. We use
AdamW~\citep{adam} with
$(\beta_1,\beta_2)=(0.9,0.95)$,
$\epsilon_{\mathrm{Adam}}=10^{-8}$, a learning rate of $10^{-6}$, weight decay
of $0.1$, cosine scheduling without warmup, and gradient-norm clipping at
$1.0$. We use symmetric policy clipping with
$(\epsilon_{\mathrm{low}},\epsilon_{\mathrm{high}})=(0.20,0.20)$, no hard
clipping of advantages, and token-level importance-ratio truncation at
$\omega_{\max}=2$. Rollouts use
temperature $0.7$, top-$p$ $1.0$, top-$k$ $20$, and a presence penalty of
$1.5$. Actor-only training takes approximately $2$ hours per epoch. For the
full MR-IQA-2 framework, four GPUs are allocated to Actor training and four to
Editor--Judge inference, increasing the per-epoch time to approximately $12$
hours.

\subsection{Rating Performance}
\label{sec:rating-performance}

\paragraph{Compared methods.}
Table~\ref{tab:main} compares MR-IQA-2 with representative BIQA methods across
six benchmarks. The hand-crafted group includes NIQE~\citep{NIQE} and
BRISQUE~\citep{BRISQUE}. Deep-learning-based methods include
NIMA~\citep{nima}, DBCNN~\citep{dbcnn}, MUSIQ~\citep{musiq},
MANIQA~\citep{yang2022maniqa}, and CLIP-IQA+~\citep{wang2023exploring}.
Among MLLM-based methods, SFT approaches include
C2Score~\citep{zhu2024adaptive}, Q-Align~\citep{q-align}, and
DeQA~\citep{deqa}, while RL approaches include Q-Insight~\citep{q-insight},
VQ-R1~\citep{Visualquality-r1}, and MR-IQA~\citep{mr-iqa}. We further compare
with the tool-augmented RL method Zoom-IQA~\citep{zoom-iqa}. For MR-IQA-2, we
report an active-vision checkpoint and a frozen-vision variant, denoted by
$^{*}$.

\paragraph{Efficiency and performance.}
MR-IQA-2 is designed primarily to improve reasoning faithfulness rather than
maximize rating. Nevertheless, it achieves competitive rating
performance. \textbf{(1)} Zoom-IQA~\citep{zoom-iqa} uses a two-stage SFT--RL
pipeline with additional annotations, whereas MR-IQA-2 uses only RL training
without additional data. The frozen-vision variant reaches an average
PLCC/SRCC of 0.840/0.816, comparable to Zoom-IQA's 0.840/0.819.
\textbf{(2)} Freezing the vision encoder and aligner is particularly effective
on AI-generated and synthetic datasets. The frozen variant achieves the best
CSIQ PLCC/SRCC of 0.840/0.801, whereas active visual training yields its main
gains on the authentic datasets.

\subsection{Faithful Reasoning Performance}
\label{sec:faithful-reasoning-performance}
\paragraph{Quality Gain as a Proxy for Reasoning Faithfulness.}
Prior work, including Q-Insight~\citep{q-insight} and
VisualQuality-R1~\citep{Visualquality-r1}, typically evaluates reasoning through human inspection of a small number of examples or indirectly through improvements in rating performance. These practices lack a unified evaluation criterion. In this work, we instead use the image-quality gain produced by reasoning-guided editing as the evaluation criterion (Eq.~(\ref{eq:reasoning-reward})), rather than rating performance. We assume that faithful reasoning should identify quality-limiting factors whose correction improves the given image.
\paragraph{Cross-editor/judge stability.}
Table~\ref{tab:cross-tool-reasoning} examines whether the Actor overfits to the
frozen Editor--Judge during training. We first replace the FLUX.2 [klein] Editor~\citep{flux2} with
Mage-Flow~\citep{zhang2026mageflow}. MR-IQA-2 retains larger quality gains than
the competing Actors, indicating that its reasoning is not specific to a single
Editor. We then replace the pretrained Judge with
Q-Insight~\citep{q-insight} and GPT-5.6 Sol~\citep{openai2026gpt56sol}. These
alternative Judges preserve the direction of the measured gains. These results indicate that the Actor learns
generalizable quality knowledge rather than exploiting idiosyncrasies of the
training-time Editor or Judge.

\paragraph{Quality and behavior analysis.}
Additional analyses (in Appendix Figure 4) provide four observations. (a) Across all datasets,
lower-quality images obtain larger quality gains. (b) Changes in sharpness are
strongly associated with Judge preference. (c) Higher-quality images tend to
remain closer to their originals after editing. (d) Training shifts the Actor
toward quality-relevant reasoning tokens.

\subsection{Ablation Study}
\label{sec:ablation-study}

\paragraph{Evaluation.}
Table~\ref{tab:credit-ablation-rating} reports ablations on rating alignment and
reasoning-guided quality improvement. Rating-only training already achieves
competitive rating alignment, with an average PLCC/SRCC of $0.836/0.815$. However,
its mean quality gain is $-0.260$, below the baseline result of $-0.136$.
Rating performance alone therefore does not guarantee faithful reasoning.
At E5, the variants without and with the credit mask reach average PLCC/SRCC
values of $0.834/0.812$ and $0.838/0.812$, respectively. Both produce positive
quality gains. The variant without the mask reaches a larger raw gain of
$+1.442$, but collapses to one normalized solution. The credit-mask variant
retains 23,457 normalized solutions while achieving a gain of $+1.079$.
Therefore, raw gain alone can favor a universal edit and does not establish
image-conditioned reasoning. The final E5 comparison is diagnostic because
the two variants also differ in KL scope; controlled E1 results are reported in
Appendix~\ref{app:credit-mask-audit}.

The credit mask is more beneficial early in training. At step 30 on the
200-image validation set, it reaches a PLCC/SRCC of $0.745/0.772$, compared
with $0.667/0.617$ without the credit mask and $0.616/0.610$ for the baseline.
Its quality gain is lower at this stage, at $0.255$ versus $0.387$ without the
credit mask. This suggests that the credit mask accelerates rating convergence
but may temporarily limit reasoning improvement. Once rating alignment
stabilizes, its benefit becomes smaller. This behavior warrants further study
of the imbalance between rating and reasoning rewards.

\begin{table*}[t]
\centering
{\small
\setlength{\tabcolsep}{2pt}
\begin{tabular}{@{}l cc cc cc cc cc cc cc@{}}
\toprule
\multicolumn{15}{@{}l}{\textit{Rating alignment}} \\
& \multicolumn{6}{c}{Authentic} & \multicolumn{2}{c}{AI-generated} &
\multicolumn{4}{c}{Synthetic} & \multicolumn{2}{c}{Average} \\
\cmidrule(lr){2-7} \cmidrule(lr){8-9} \cmidrule(lr){10-13}
\cmidrule(lr){14-15}
& \multicolumn{2}{c}{KonIQ} & \multicolumn{2}{c}{SPAQ} &
\multicolumn{2}{c}{LIVE-W} & \multicolumn{2}{c}{AGIQA-3K} &
\multicolumn{2}{c}{KADID-10k} & \multicolumn{2}{c}{CSIQ} &
\multicolumn{2}{c}{Avg.} \\
\cmidrule(lr){2-3} \cmidrule(lr){4-5} \cmidrule(lr){6-7}
\cmidrule(lr){8-9} \cmidrule(lr){10-11} \cmidrule(lr){12-13}
\cmidrule(lr){14-15}
\textbf{Method} & PLCC & SRCC & PLCC & SRCC & PLCC & SRCC & PLCC & SRCC &
PLCC & SRCC & PLCC & SRCC & PLCC & SRCC \\
\midrule
Baseline
& 0.553 & 0.583 & 0.750 & 0.743 & 0.592 & 0.564
& 0.679 & 0.655 & 0.660 & 0.669 & 0.681 & 0.657 & 0.652 & 0.645 \\
Rating-only
& \textbf{0.952} & \textbf{0.938} & \textbf{0.901} & 0.899
& \textbf{0.896} & \textbf{0.875}
& 0.808 & \textbf{0.740} & 0.669 & 0.676 & 0.789 & 0.764
& 0.836 & \textbf{0.815} \\
Without credit mask (E5)
& 0.932 & 0.914 & 0.896 & 0.897 & 0.878 & 0.857
& 0.809 & 0.736 & \textbf{0.676} & \textbf{0.684}
& 0.815 & \textbf{0.786} & 0.834 & 0.812 \\
Credit mask (E5)
& 0.937 & 0.917 & 0.900 & \textbf{0.899} & 0.893 & 0.863
& \textbf{0.809} & 0.739 & 0.667 & 0.669
& \textbf{0.824} & 0.785 & \textbf{0.838} & 0.812 \\
\midrule
\multicolumn{15}{@{}l}{\textit{Subset quality gains}} \\
\textbf{Method}
& \multicolumn{3}{c}{\textbf{Original mean}}
& \multicolumn{4}{c}{\textbf{Edited mean}}
& \multicolumn{3}{c}{\textbf{Mean gain}}
& \multicolumn{4}{c}{\textbf{Positive gain (\%)}} \\
\cmidrule(lr){2-4} \cmidrule(lr){5-8} \cmidrule(lr){9-11}
\cmidrule(lr){12-15}
Baseline
& \multicolumn{3}{c}{$2.844$}
& \multicolumn{4}{c}{$2.708$}
& \multicolumn{3}{c}{$-0.136$}
& \multicolumn{4}{c}{$40.664\%$} \\
Rating-only
& \multicolumn{3}{c}{$2.844$}
& \multicolumn{4}{c}{$2.584$}
& \multicolumn{3}{c}{$-0.260$}
& \multicolumn{4}{c}{$30.686\%$} \\
Without credit mask (E5)
& \multicolumn{3}{c}{$2.844$}
& \multicolumn{4}{c}{$\mathbf{4.286}$}
& \multicolumn{3}{c}{$\mathbf{+1.442}$}
& \multicolumn{4}{c}{$\mathbf{99.997\%}$} \\
Credit mask (E5)
& \multicolumn{3}{c}{$2.844$}
& \multicolumn{4}{c}{$3.923$}
& \multicolumn{3}{c}{$+1.079$}
& \multicolumn{4}{c}{$98.813\%$} \\
\bottomrule
\end{tabular}
}
\caption{\textbf{Rating and quality-gain ablations.} The upper panel reports
PLCC/SRCC on the complete test sets and their six-dataset average. The
lower panel reports Judge scores before and after reasoning-guided editing.
Baseline and Rating-only use 5,843 common images; the E5 rows use their full
successful audits (28,270 without the mask and 28,044 with the mask). Bold
marks the best scalar result before rounding. The higher unmasked gain
coincides with single-solution collapse and is not interpreted as improved
reasoning.}
\label{tab:credit-ablation-rating}
\end{table*}

\begin{figure*}[!t]
\centering
\includegraphics[width=\textwidth]{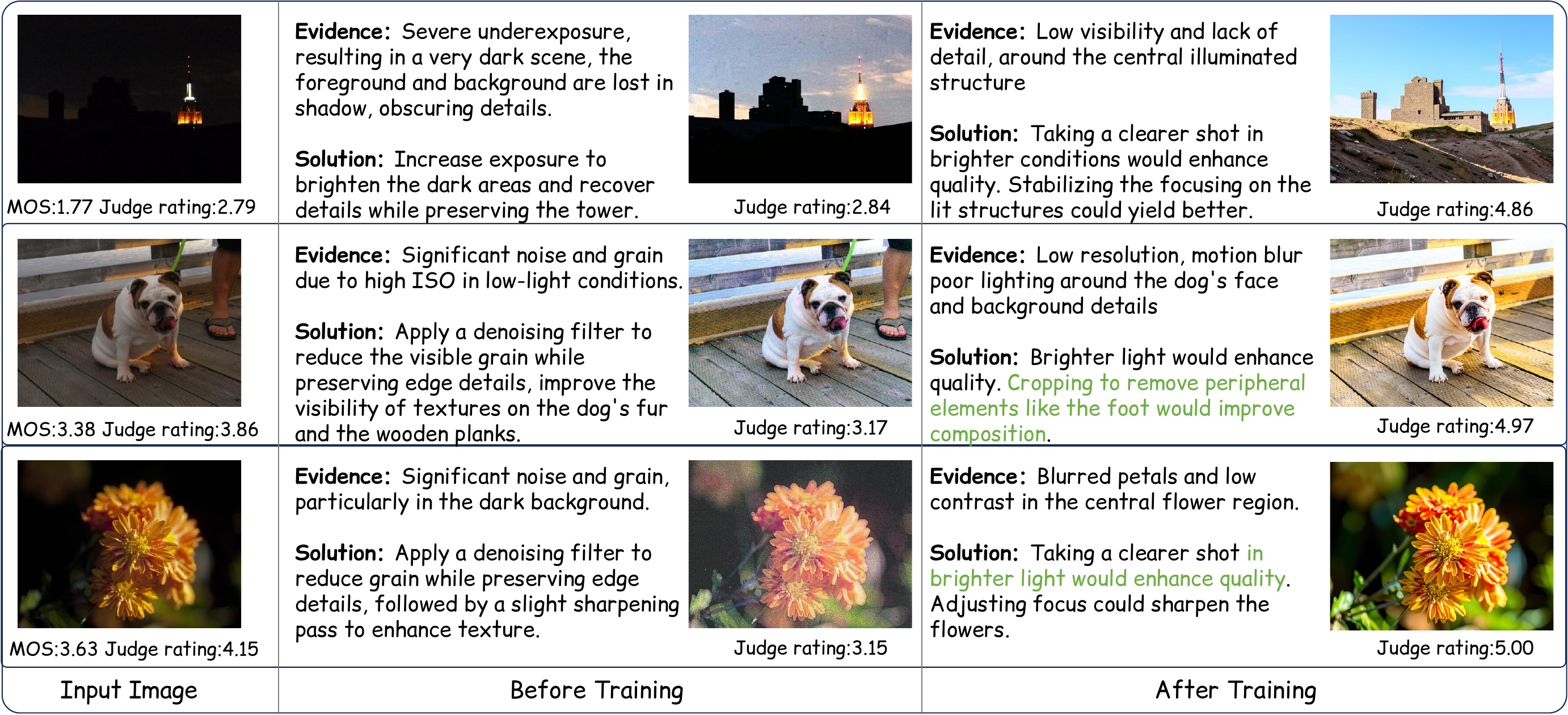}
\caption{\textbf{Qualitative comparison before and after training.} Each row
shows the input image and the Actor's reasoning-guided edits before and after
training. Post-training reasoning yields more effective interventions and
larger Judge-rated quality gains. Green text highlights additional quality
factors identified after training.}
\label{fig:case-study}
\end{figure*}

\subsection{Case Study}
\label{sec:case-study}
Figure~\ref{fig:case-study} compares reasoning-guided edits before and after
training. After training, all three edits receive higher Judge ratings, and the
Actor proposes more specific interventions. In the dog example, it recommends
cropping the distracting foot near the image boundary to improve visual appeal.
The examples also expose a semantic-preservation problem: an edited image may
no longer retain the meaning of the original. In the first row, the Editor
changes a night sky into a blue daytime sky. The boundary between quality
enhancement and semantic alteration therefore requires further study.

\section{Discussion}
\label{sec:discussion}
\paragraph{Good Evidence, Bad Solution?}
In this work, we jointly treat \texttt{evidence} and \texttt{solution} as reasoning. However, as with human perception, a model may correctly identify why an image has poor quality yet fail to propose an effective correction. Our training necessarily encourages the model to discover effective solutions through exploration, raising the question of whether this process also improves its evidence. Maybe the evidence capability does not improve. But we argue that a solution that improves image quality through relevant visual changes indicates an improved understanding of the underlying quality factors. Explicitly modeling the causal relation between evidence and solution may provide a more reliable direction for future work.

\section{Limitations and Future Work}
\label{sec:limitations}
This work has two main limitations. First, the current framework does not yet
realize fully closed-loop visual reflection. The edited image is evaluated by
the Judge but is not returned to the Actor for further visual reasoning.
Second, using separate Actor, Editor, and Judge models introduces computational
and parameter redundancy.
Future work will close the reflection loop by feeding the edited image back to
the Actor, enabling direct comparison between the original and edited images.
We will also explore smaller models and parameter sharing across modules to
improve efficiency.
\section{Conclusion}
\label{sec:conclusion}
In this work, we aim to supervise the faithfulness of image-quality reasoning,
moving BIQA beyond plausible explanations toward visually verifiable
understanding. We use quality gain as an operational measure of reasoning
faithfulness and propose MR-IQA-2, an Actor--Editor--Judge framework. The
Actor's reasoning guides the Editor, while the Judge evaluates the resulting
quality change to test whether the identified factors are supported.
Fine-grained credit assignment further decouples reasoning and rating
supervision to reduce reward ambiguity. MR-IQA-2 retains competitive rating
alignment while producing more reliable and actionable quality reasoning.
More broadly, our framework provides a new perspective on evaluating reasoning
faithfulness, links blind assessment to reference-based verification through
intervention-generated image pairs, and suggests a path toward modeling human
visual preferences and aesthetic judgments.

\bibliography{aaai2027}

@String(AAAI = {AAAI})

@inproceedings{yu2025dapo,
  title={{DAPO}: An Open-Source {LLM} Reinforcement Learning System at Scale},
  author={Yu, Qiying and Zhang, Zheng and Zhu, Ruofei and Yuan, Yufeng and Zuo, Xiaochen and others},
  booktitle={Advances in Neural Information Processing Systems},
  volume={38},
  year={2025}
}

@inproceedings{q-instruct,
  title={{Q-Instruct}: Improving Low-Level Visual Abilities for Multi-Modality Foundation Models},
  author={Wu, Haoning and Zhang, Zicheng and Zhang, Erli and Chen, Chaofeng and Liao, Liang and Wang, Annan and Xu, Kaixin and Li, Chunyi and Hou, Jingwen and Zhai, Guangtao and others},
  booktitle={Proceedings of the IEEE/CVF conference on computer vision and pattern recognition},
  pages={25490--25500},
  year={2024}
}

@article{q-insight,
  title={{Q-Insight}: Understanding Image Quality via Visual Reinforcement Learning},
  author={Li, Weiqi and Zhang, Xuanyu and Zhao, Shijie and Zhang, Yabin and Li, Junlin and Zhang, Li and Zhang, Jian},
  journal={arXiv preprint arXiv:2503.22679},
  year={2025}
}

@article{dbcnn,
title={Blind Image Quality Assessment Using A Deep Bilinear Convolutional Neural Network},
author={Zhang, Weixia and Ma, Kede and Yan, Jia and Deng, Dexiang and Wang, Zhou},
journal={IEEE Transactions on Circuits and Systems for Video Technology},
volume={30},
number={1},
pages={36--47},
year={2020}
}

@inproceedings{musiq,
  title={MUSIQ: Multi-scale Image Quality Transformer},
  author={Ke, Junjie and Wang, Qifei and Wang, Yilin and Milanfar, Peyman and Yang, Feng},
  booktitle={Proceedings of the IEEE/CVF International Conference on Computer Vision},
  pages={5148--5157},
  year={2021}
}

@inproceedings{arniqa,
  title={ARNIQA: Learning Distortion Manifold for Image Quality Assessment},
  author={Agnolucci, Lorenzo and Galteri, Leonardo and Bertini, Marco and Del Bimbo, Alberto},
  booktitle={Proceedings of the IEEE/CVF Winter Conference on Applications of Computer Vision},
  pages={189--198},
  year={2024}
}

@article{topiq,
  title={{TOPIQ}: A Top-Down Approach from Semantics to Distortions for Image Quality Assessment},
  author={Chen, Chaofeng and Mo, Jiadi and Hou, Jingwen and Wu, Haoning and Liao, Liang and Sun, Wenxiu and Yan, Qiong and Lin, Weisi},
  journal={IEEE Transactions on Image Processing},
  volume={33},
  pages={2404--2418},
  year={2024},
  publisher={IEEE}
}

@article{qwen2.5,
  title={{Qwen2.5-VL} Technical Report},
  author={Bai, Shuai and Chen, Keqin and Liu, Xuejing and Wang, Jialin and Ge, Wenbin and Song, Sibo and Dang, Kai and Wang, Peng and Wang, Shijie and Tang, Jun and others},
  journal={arXiv preprint arXiv:2502.13923},
  year={2025}
}

@inproceedings{liqe,
  title={Blind image quality assessment via vision-language correspondence: A multitask learning perspective},
  author={Zhang, Weixia and Zhai, Guangtao and Wei, Ying and Yang, Xiaokang and Ma, Kede},
  booktitle={Proceedings of the IEEE/CVF conference on computer vision and pattern recognition},
  pages={14071--14081},
  year={2023}
}

@inproceedings{depictqa,
  title={Depicting beyond scores: Advancing image quality assessment through multi-modal language models},
  author={You, Zhiyuan and Li, Zheyuan and Gu, Jinjin and Yin, Zhenfei and Xue, Tianfan and Dong, Chao},
  booktitle={European Conference on Computer Vision},
  pages={259--276},
  year={2024},
  organization={Springer}
}

@article{q-align,
  title={Q-align: Teaching lmms for visual scoring via discrete text-defined levels},
  author={Wu, Haoning and Zhang, Zicheng and Zhang, Weixia and Chen, Chaofeng and Liao, Liang and Li, Chunyi and Gao, Yixuan and Wang, Annan and Zhang, Erli and Sun, Wenxiu and others},
  journal={arXiv preprint arXiv:2312.17090},
  year={2023}
}

@article{csiq,
  title={Most apparent distortion: full-reference image quality assessment and the role of strategy},
  author={Larson, Eric C and Chandler, Damon M},
  journal={Journal of electronic imaging},
  volume={19},
  number={1},
  pages={011006--011006},
  year={2010},
  publisher={Society of Photo-Optical Instrumentation Engineers}
}

@inproceedings{spaq,
  title={Perceptual quality assessment of smartphone photography},
  author={Fang, Yuming and Zhu, Hanwei and Zeng, Yan and Ma, Kede and Wang, Zhou},
  booktitle={Proceedings of the IEEE/CVF conference on computer vision and pattern recognition},
  pages={3677--3686},
  year={2020}
}

@article{koniq,
  title={{KonIQ-10k}: An Ecologically Valid Database for Deep Learning of Blind Image Quality Assessment},
  author={Hosu, Vlad and Lin, Hanhe and Sziranyi, Tamas and Saupe, Dietmar},
  journal={IEEE Transactions on Image Processing},
  volume={29},
  pages={4041--4056},
  year={2020},
  publisher={IEEE}
}

@inproceedings{kadid,
  title={KADID-10k: A large-scale artificially distorted IQA database},
  author={Lin, Hanhe and Hosu, Vlad and Saupe, Dietmar},
  booktitle={2019 Eleventh International Conference on Quality of Multimedia Experience (QoMEX)},
  pages={1--3},
  year={2019},
  organization={IEEE}
}

@article{agiqa,
  title={{AGIQA-3K}: An Open Database for AI-Generated Image Quality Assessment},
  author={Li, Chunyi and Zhang, Zicheng and Wu, Haoning and Sun, Wei and Min, Xiongkuo and Liu, Xiaohong and Zhai, Guangtao and Lin, Weisi},
  journal={IEEE Transactions on Circuits and Systems for Video Technology},
  volume={34},
  number={8},
  pages={6833--6846},
  year={2024},
  publisher={IEEE}
}

@article{live-w,
  title={Live in the wild image quality challenge database},
  author={Ghadiyaram, Deepti and Bovik, Alan C},
  journal={Online: http://live. ece. utexas. edu/research/ChallengeDB/index. html [Mar, 2017]},
  year={2015}
}

@inproceedings{deqa,
  title={Teaching large language models to regress accurate image quality scores using score distribution},
  author={You, Zhiyuan and Cai, Xin and Gu, Jinjin and Xue, Tianfan and Dong, Chao},
  booktitle={Proceedings of the Computer Vision and Pattern Recognition Conference},
  pages={14483--14494},
  year={2025}
}

@article{NIQE,
  title={Making a “completely blind” image quality analyzer},
  author={Mittal, Anish and Soundararajan, Rajiv and Bovik, Alan C},
  journal={IEEE Signal processing letters},
  volume={20},
  number={3},
  pages={209--212},
  year={2012},
  publisher={IEEE}
}

@article{BRISQUE,
  title={No-reference image quality assessment in the spatial domain},
  author={Mittal, Anish and Moorthy, Anush Krishna and Bovik, Alan Conrad},
  journal={IEEE Transactions on image processing},
  volume={21},
  number={12},
  pages={4695--4708},
  year={2012},
  publisher={IEEE}
}

@article{nima,
  title={NIMA: Neural image assessment},
  author={Talebi, Hossein and Milanfar, Peyman},
  journal={IEEE transactions on image processing},
  volume={27},
  number={8},
  pages={3998--4011},
  year={2018},
  publisher={IEEE}
}

@inproceedings{wang2023exploring,
  title={Exploring clip for assessing the look and feel of images},
  author={Wang, Jianyi and Chan, Kelvin CK and Loy, Chen Change},
  booktitle={Proceedings of the AAAI conference on artificial intelligence},
  volume={37},
  pages={2555--2563},
  year={2023}
}

@inproceedings{yang2022maniqa,
  title={Maniqa: Multi-dimension attention network for no-reference image quality assessment},
  author={Yang, Sidi and Wu, Tianhe and Shi, Shuwei and Lao, Shanshan and Gong, Yuan and Cao, Mingdeng and Wang, Jiahao and Yang, Yujiu},
  booktitle={Proceedings of the IEEE/CVF conference on computer vision and pattern recognition},
  pages={1191--1200},
  year={2022}
}

@article{zhu2024adaptive,
  title={Adaptive image quality assessment via teaching large multimodal model to compare},
  author={Zhu, Hanwei and Wu, Haoning and Li, Yixuan and Zhang, Zicheng and Chen, Baoliang and Zhu, Lingyu and Fang, Yuming and Zhai, Guangtao and Lin, Weisi and Wang, Shiqi},
  journal={Advances in Neural Information Processing Systems},
  volume={37},
  pages={32611--32629},
  year={2024}
}

@inproceedings{adam,
  title={Decoupled Weight Decay Regularization},
  author={Loshchilov, Ilya and Hutter, Frank},
  booktitle={International Conference on Learning Representations},
  year={2019},
  url={https://openreview.net/forum?id=Bkg6RiCqY7}
}

@inproceedings{briqa,
  title={Building Reasonable Inference for Vision-Language Models in Blind Image Quality Assessment},
  author={Li, Yuan and Sun, Zitang and Chen, Yen-ju and Nishida, Shin’ya},
  booktitle={International Conference on Neural Information Processing},
  pages={283--295},
  year={2025},
  organization={Springer}
}

@article{h-iqa,
  title={Guiding Perception-Reasoning Closer to Human in Blind Image Quality Assessment},
  author={Li, Yuan and Yu, Yahan and Lin, Youyuan and Yang, Yong-Hao and Chu, Chenhui and Nishida, Shin'ya},
  journal={arXiv preprint arXiv:2512.16484},
  year={2025}
}

@article{zoom-iqa,
  title={{Zoom-IQA}: Image Quality Assessment with Reliable Region-Aware Reasoning},
  author={Liang, Guoqiang and Wang, Jianyi and Wu, Zhonghua and Zhou, Shangchen and Loy, Chen Change},
  journal={arXiv preprint arXiv:2601.02918},
  year={2026}
}

@article{tool-iqa,
  title={{Tool-IQA}: Augmenting Image Quality Assessment with Simple Tools},
  author={Qin, Guanyi and Zhang, Junjie and He, Chunming and Fu, Yibing and Liang, Jie and Wu, Tianhe and Zhang, Lei},
  journal={arXiv preprint arXiv:2606.16082},
  year={2026}
}

@article{Visualquality-r1,
  title={{VisualQuality-R1}: Reasoning-Induced Image Quality Assessment via Reinforcement Learning to Rank},
  author={Wu, Tianhe and Zou, Jian and Liang, Jie and Zhang, Lei and Ma, Kede},
  journal={Advances in Neural Information Processing Systems},
  volume={38},
  pages={88167--88190},
  year={2025}
}

@inproceedings{wang2003msssim,
  title={Multi-Scale Structural Similarity for Image Quality Assessment},
  author={Wang, Zhou and Simoncelli, Eero P. and Bovik, Alan C.},
  booktitle={Proceedings of the 37th Asilomar Conference on Signals, Systems and Computers},
  volume={2},
  pages={1398--1402},
  year={2003},
  doi={10.1109/ACSSC.2003.1292216}
}

@inproceedings{zhang2018lpips,
  title={The Unreasonable Effectiveness of Deep Features as a Perceptual Metric},
  author={Zhang, Richard and Isola, Phillip and Efros, Alexei A. and Shechtman, Eli and Wang, Oliver},
  booktitle={Proceedings of the IEEE Conference on Computer Vision and Pattern Recognition},
  pages={586--595},
  year={2018}
}

@article{szekely2007distance,
  title={Measuring and Testing Dependence by Correlation of Distances},
  author={Sz{\'e}kely, G{\'a}bor J. and Rizzo, Maria L. and Bakirov, Nail K.},
  journal={The Annals of Statistics},
  volume={35},
  number={6},
  pages={2769--2794},
  year={2007},
  doi={10.1214/009053607000000505}
}

@article{qwen3vl,
  title={Qwen3-vl technical report},
  author={Bai, Shuai and Cai, Yuxuan and Chen, Ruizhe and Chen, Keqin and Chen, Xionghui and Cheng, Zesen and Deng, Lianghao and Ding, Wei and Gao, Chang and Ge, Chunjiang and others},
  journal={arXiv preprint arXiv:2511.21631},
  year={2025}
}

@article{shao2024deepseekmath,
  title={Deepseekmath: Pushing the limits of mathematical reasoning in open language models},
  author={Shao, Zhihong and Wang, Peiyi and Zhu, Qihao and Xu, Runxin and Song, Junxiao and Bi, Xiao and Zhang, Haowei and Zhang, Mingchuan and Li, YK and Wu, Yang and others},
  journal={arXiv preprint arXiv:2402.03300},
  year={2024}
}

@misc{mr-iqa,
      title={MR-IQA: A Unified Margin View of Regression and Ranking for Blind Image Quality Assessment},
      author={Yuan Li and Youyuan Lin and Zitang Sun and Yung-Hao Yang and Kiyofumi Miyoshi and Chenhui Chu and Shin'ya Nishida},
      year={2026},
      eprint={2606.29760},
      archivePrefix={arXiv},
      primaryClass={cs.CV},
      url={https://arxiv.org/abs/2606.29760},
}

@misc{qwen35,
  title={{Qwen3.5}: Towards Native Multimodal Agents},
  author={{Qwen Team}},
  month={February},
  year={2026},
  url={https://qwen.ai/blog?id=qwen3.5}
}

@misc{flux2,
  author={{Black Forest Labs}},
  title={{FLUX.2}: Frontier Visual Intelligence},
  year={2025},
  howpublished={\url{https://bfl.ai/blog/flux-2}}
}

@article{zhang2026mageflow,
  title={Mage-Flow: An Efficient Native-Resolution Foundation Model for Image Generation and Editing},
  author={Zhang, Xinjie and Zhang, Peng and Zheng, Shicheng and Guo, Jinghao and Jia, Zhaoyang and Shen, Yifei and Guo, Xun and Luo, Yuxuan and Li, Jiahao and Xie, Wenxuan and Pu, Fanyi and Zhang, Xiaoyi and Zhang, Kaichen and Guo, Zongyu and Bi, Tianci and Gui, Dongnan and Liu, Zhening and Wen, Zimo and Zheng, Zihan and Yang, Senqiao and Li, Xiao and Wang, Jinglu and Li, Bin and Lu, Yan},
  journal={arXiv preprint arXiv:2607.19064},
  year={2026}
}

@misc{openai2026gpt56sol,
  author={{OpenAI}},
  title={{GPT-5.6 Sol Model}},
  year={2026},
  howpublished={\url{https://developers.openai.com/api/docs/models/gpt-5.6-sol}}
}

\clearpage
\appendix

In this appendix, we provide technical details and more comprehensive analyses
of our proposed method. It comprises Section~\ref{app:backbone},
\emph{Backbone, Algorithm and Hyperparameters};
Section~\ref{app:visual-quality-feature-analysis},
\emph{Visual Quality Feature Analysis}; Section~\ref{app:editor-proxy},
\emph{Discussion on the Editor Proxy};
Section~\ref{app:joint-rating-reasoning},
\emph{Joint Rating, Reasoning, and Diversity Analysis}; and
Section~\ref{app:credit-mask-audit}, \emph{Detailed Credit-Mask Audit}.
\section{Backbone, Algorithm and Hyperparameters}
\label{app:backbone}

\subsection{Dataset Protocol.}
Our training and evaluation settings follow the same configuration as the main
paper. We train on the KonIQ-10k training subset~\citep{koniq} and use a fixed
set of 200 images randomly sampled from the KonIQ-10k test split to monitor
early-stage convergence. Final testing is conducted on the KonIQ-10k test
split, SPAQ~\citep{spaq}, LIVE-W~\citep{live-w},
AGIQA-3K~\citep{agiqa}, KADID-10k~\citep{kadid}, and CSIQ~\citep{csiq}.

At the time of manuscript completion, we did not have access to
Zoom-IQA~\citep{zoom-iqa} and therefore could not reproduce it under our
protocol. Tool-IQA~\citep{tool-iqa} follows a different training--testing
configuration; consequently, we cannot report directly comparable Tool-IQA
results for some experiments.

For the cross Actor--Judge--Editor evaluation reported in Table~2 of the main
paper, we use a fixed 5,866-image subset, randomly sampling 1,000 images from
each test set except CSIQ, for which all 866 images are used. This subsampling
is necessary for computational efficiency: evaluating the full test sets would
involve approximately 23,000 images, each requiring four editing operations
and three Judge evaluations, resulting in a prohibitively heavy inference
cost.

\subsection{Candidate Backbones.}
The Qwen series has been widely adopted in recent BIQA studies due to its open-source availability, strong multimodal capability, and computational efficiency.
For example, existing methods such as Q-Insight~\citep{q-insight},
VQ-R1~\citep{Visualquality-r1}, and Zoom-IQA~\citep{zoom-iqa} employ
Qwen2.5-VL-7B~\citep{qwen2.5} as their backbone model.
However, despite its strong performance, the 7B-scale model still introduces considerable computational overhead, which limits its practical deployment and training efficiency.

Therefore, in this work, we investigate more lightweight backbone configurations with 2B and 4B parameters.
Considering future scalability and the continuous evolution of multimodal large language models, we further evaluate the latest Qwen3-VL and Qwen3.5-VL architectures.
We conduct comprehensive backbone comparisons among 2B/4B variants of Qwen3 and Qwen3.5 to identify an optimal trade-off between training efficiency and performance.

\paragraph{Qwen3-VL versus Qwen3.5.}
The T1--T2 comparison in Table~\ref{tab:preliminary-ablation} shows that, under
matched frozen-vision DAPO settings, Qwen3.5-2B outperforms Qwen3-VL-2B on five
of six datasets for both PLCC and SRCC, and on all six in MAE. It also requires
only 0.81$\times$ the training time per epoch.

\paragraph{2B versus 4B.}
The T6--T7 comparison in Table~\ref{tab:preliminary-ablation} shows that the 4B
model converges faster in rating performance. By Epoch 2, T7 reaches a
validation PLCC/SRCC/MAE of
0.929/0.919/0.566, compared with 0.915/0.904/0.744 for the 2B T6; by Epoch 3,
T7 further improves to 0.935/0.924/0.228. The 4B model also produces more
diverse outputs, with 83 distinct ratings, 200 unique completions, and a
41.5\% U-score, compared with a 39.0\% U-score for T6. We further observe that
its response lengths are better aligned with our objective of eliciting
detailed quality reasoning.

\begin{table*}[!t]
\centering
{\small
\setlength{\tabcolsep}{4.5pt}
\begin{tabular}{@{}lcrccr@{}}
\toprule
\textbf{Variant} & \textbf{Iter.} & \textbf{Time (h/ep.)} &
\textbf{Val. P/S/M} & \textbf{U-score (\%)} & \textbf{Gen. P/S/M} \\
\midrule
\multicolumn{6}{@{}l}{\emph{(a) Qwen3-VL-2B}} \\
T1: DAPO i1, frozen, global & 1 & 1.38 & 0.831/0.817/0.868 & 12.0 & 0.809/0.787/0.936 \\
\addlinespace[1pt]
\multicolumn{6}{@{}l}{\emph{(b) Qwen3.5-2B}} \\
T2: DAPO i1, frozen, global & 1 & 1.13 & 0.890/0.881/0.691 & 36.0 & \textbf{0.835}/0.808/0.740 \\
T3: DAPO i4, frozen, global & 4 & 2.47 & 0.910/0.904/0.265 & 21.5 & 0.834/0.809/0.541 \\
T4: GRPO i1, frozen & 1 & 0.81 & 0.886/0.883/0.845 & 34.0 & 0.829/0.800/0.878 \\
T5: DAPO i1, visual, global & 1 & 1.30 & \textbf{0.929}/\textbf{0.921}/\textbf{0.261} & 35.0 & 0.828/0.809/\textbf{0.513} \\
T6: DAPO i1, visual, local-six & 1 & 1.54 & 0.917/0.906/0.587 & 39.0 & \textbf{0.835}/\textbf{0.813}/0.684 \\
\addlinespace[1pt]
\multicolumn{6}{@{}l}{\emph{(c) Qwen3.5-4B, local-six}} \\
T7: DAPO i1, visual, local-six & 1 & 2.56 & 0.935/0.924/0.228 & 41.5 & -- \\
\bottomrule
\end{tabular}
}
\caption{Actor-only ablations at their reported endpoints.
$\mathrm{P}/\mathrm{S}/\mathrm{M}$ denotes PLCC/SRCC/MAE; Val. is the aligned
200-image set, and Gen. is the six-dataset average. U-score measures
distinct validation ratings, and Time is the mean wall-clock time per epoch.
Bold marks the best alignment result among the controlled Qwen3.5-2B variants;
higher PLCC/SRCC and lower MAE are better. All variants use E3.}
\label{tab:preliminary-ablation}
\end{table*}

\begin{figure*}[!t]
\centering
\includegraphics[width=\textwidth]{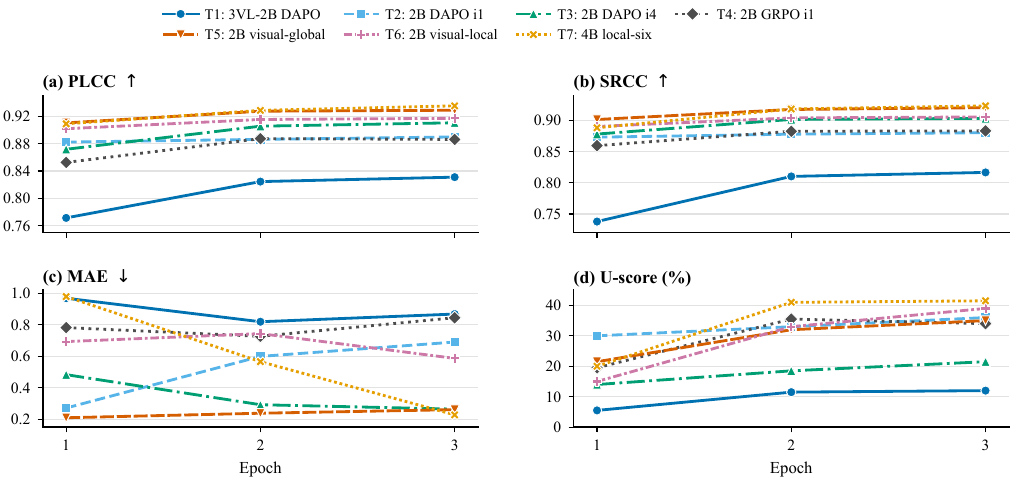}
\caption{Validation trajectories through E3 on the common 200-image set. T7
denotes the 4B local-six run. Curves are single fixed-seed runs, so no error
bands are shown. Higher PLCC/SRCC and
lower MAE are better, while U-score diagnoses rating diversity. Color, marker,
and line style consistently identify T1--T7.}
\label{fig:ablation-epochs}
\end{figure*}

\begin{figure*}[!t]
\centering
\includegraphics[width=\textwidth]{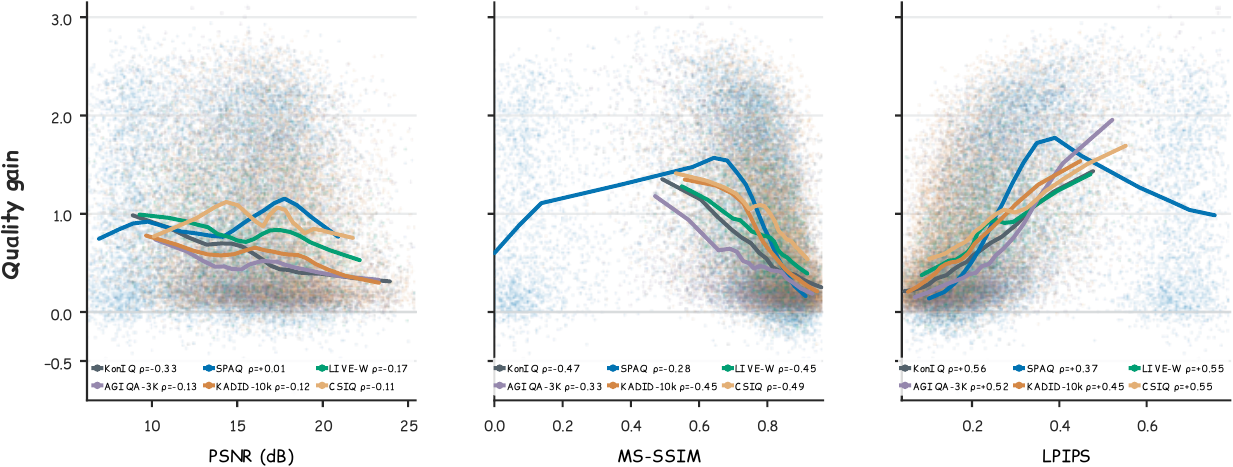}
\caption{\textbf{Quality gain versus visual change scale.}
The panels relate the Judge-observed quality gain $\Delta s$ to PSNR,
MS-SSIM~\citep{wang2003msssim}, and LPIPS~\citep{zhang2018lpips} between the
original and edited images. Colors denote datasets, curves show smoothed
trends, and $\rho$ denotes Spearman correlation.}
\label{fig:quality-fidelity}
\end{figure*}

\subsection{Optimization Algorithms: GRPO or DAPO?}
\label{app:optimization-algorithms}

Compared with GRPO~\citep{shao2024deepseekmath},
DAPO~\citep{yu2025dapo} dynamically resamples low-variance groups, thereby
increasing within-group reward diversity and preventing policy advantages from
vanishing prematurely. We therefore adopted DAPO in most of our early
experiments.

Across three epochs, 3,282 of 20,832 GRPO learner groups (15.75\%) have zero
within-group reward variance and therefore zero policy advantage. DAPO
resampling reduces the proportion of zero-advantage learner groups to 0\%.
Comparing T2 and T4 in Table~\ref{tab:preliminary-ablation}, DAPO improves
the six-dataset average PLCC/SRCC by 0.006/0.008 and reduces MAE by 0.138 at
Epoch 3, with improvements on five, six, and six datasets, respectively.

Despite its better performance, DAPO requires 1.866$\times$ more sampled
trajectories and increases the training time per epoch from 0.81 to 1.13 hours
(approximately 39.5\%). This additional training-time cost is
not affordable in our training environment, especially when scaling the model
to 4B parameters or beyond. Therefore, we use GRPO as the default optimization
algorithm in the final framework.

\subsection{Hyper-Parameter Settings.}

\paragraph{Group Size: 6 versus 48.}
MR-IQA~\citep{mr-iqa} reports its best performance with local groups of six,
where rewards are computed by comparing samples only within each six-image
group. Table~\ref{tab:preliminary-ablation} compares this local-6 setting (T6)
with a larger group of 48 (T5).
The large group achieves better in-domain validation PLCC/SRCC/MAE
(0.929/0.921/0.261 versus 0.917/0.906/0.587), whereas local-6 achieves higher
six-dataset average PLCC/SRCC (0.835/0.813 versus 0.828/0.809). Because the
local-6 setting provides stronger cross-dataset correlation, we adopt a group
size of six.

\paragraph{Vision Modules: Frozen versus Active.}
The T2--T5 comparison in Table~\ref{tab:preliminary-ablation} shows that active
visual training performs better on authentic datasets, achieving
PLCC/SRCC values of 0.952/0.938 on KonIQ and 0.897/0.877 on LIVE-W, compared
with 0.925/0.904 and 0.881/0.847 for the frozen variant. In contrast, freezing
the vision encoder and aligner performs better on AGIQA-3K and the synthetic
datasets: it achieves 0.826/0.768 on AGIQA-3K, 0.665/0.676 on KADID-10k, and
0.840/0.801 on CSIQ, compared with 0.816/0.747, 0.661/0.668, and 0.771/0.746
under active visual training. Consequently, the frozen variant yields higher
six-dataset average PLCC/SRCC (0.840/0.816 versus 0.833/0.812), indicating
stronger overall generalization.

\paragraph{Rating Diversity.}
Zoom-IQA~\citep{zoom-iqa} reports score-space collapse when using a ranking
reward, with only a 2.04\% unique-score ratio. Across T1--T7 in
Table~\ref{tab:preliminary-ablation}, the E3 U-scores remain between 12.0\% and
41.5\%, corresponding to 24--83 distinct ratings on the 200-image validation
set. Thus, we do not observe rating-diversity collapse under our current
settings.

\paragraph{Repeated Sampling Updates.}
MR-IQA~\citep{mr-iqa} reuses each rollout sample for four policy updates. We
evaluate whether these repeated updates remain effective by comparing T2 and
T3 in Table~\ref{tab:preliminary-ablation}. Four updates improve validation
PLCC/SRCC/MAE from 0.890/0.881/0.691 to
0.910/0.904/0.265. However, the six-dataset average PLCC/SRCC remain nearly
unchanged (0.835/0.808 versus 0.834/0.809), while training time increases from
1.13 to 2.47 hours per epoch. Therefore, one policy update per rollout is
sufficient under our current settings.

\section{Visual Quality Feature Analysis}
\label{app:visual-quality-feature-analysis}

In our framework, the FLUX.2 [klein] Editor~\citep{flux2} produces an edited
image conditioned on the quality reasoning for each input. The resulting
original--edited pair bridges single-image quality assessment with
reference-based quality assessment, enabling us to analyze how the visual
intervention relates to the predicted quality improvement. In this section, we
investigate three questions: (1) whether larger visual changes lead to greater
quality gains; (2) how the distributions of low-level visual features change
after editing; and (3) what behavioral preferences emerge from the overall
framework.

\subsection{Visual Change and Quality Gains}
\label{app:quality-fidelity}

\paragraph{Visual change and quality improvement.}
Figure~\ref{fig:quality-fidelity} relates the Judge-observed quality gain to
three reference-based measures of the visual change between the original and
edited images. Quality gain is generally negatively correlated with PSNR
($\rho=-0.33$ to $+0.01$) and MS-SSIM ($\rho=-0.49$ to $-0.28$), and
positively correlated with LPIPS ($\rho=0.37$ to $0.56$). Because larger
changes correspond to lower
PSNR/MS-SSIM and higher LPIPS, these consistent directions indicate that
larger editing changes are generally associated with greater image-quality
improvements. SPAQ deviates from the other datasets across multiple measures. Its
quality-gain correlation is nearly zero for PSNR ($\rho=+0.01$) and is the
weakest in magnitude for both MS-SSIM ($\rho=-0.28$) and LPIPS
($\rho=+0.37$). A similar pattern appears in rating performance: while the
rating correlations on the other datasets change during training, the SPAQ
correlation remains close to 0.90 with only minor variation. This stability
suggests that SPAQ is less sensitive to the training-induced changes observed
on the other benchmarks.

\subsection{Low-Level Visual Feature Patterns}
\label{app:quality-low-level-features}

\begin{figure*}[t!]
\centering
\includegraphics[width=\textwidth]{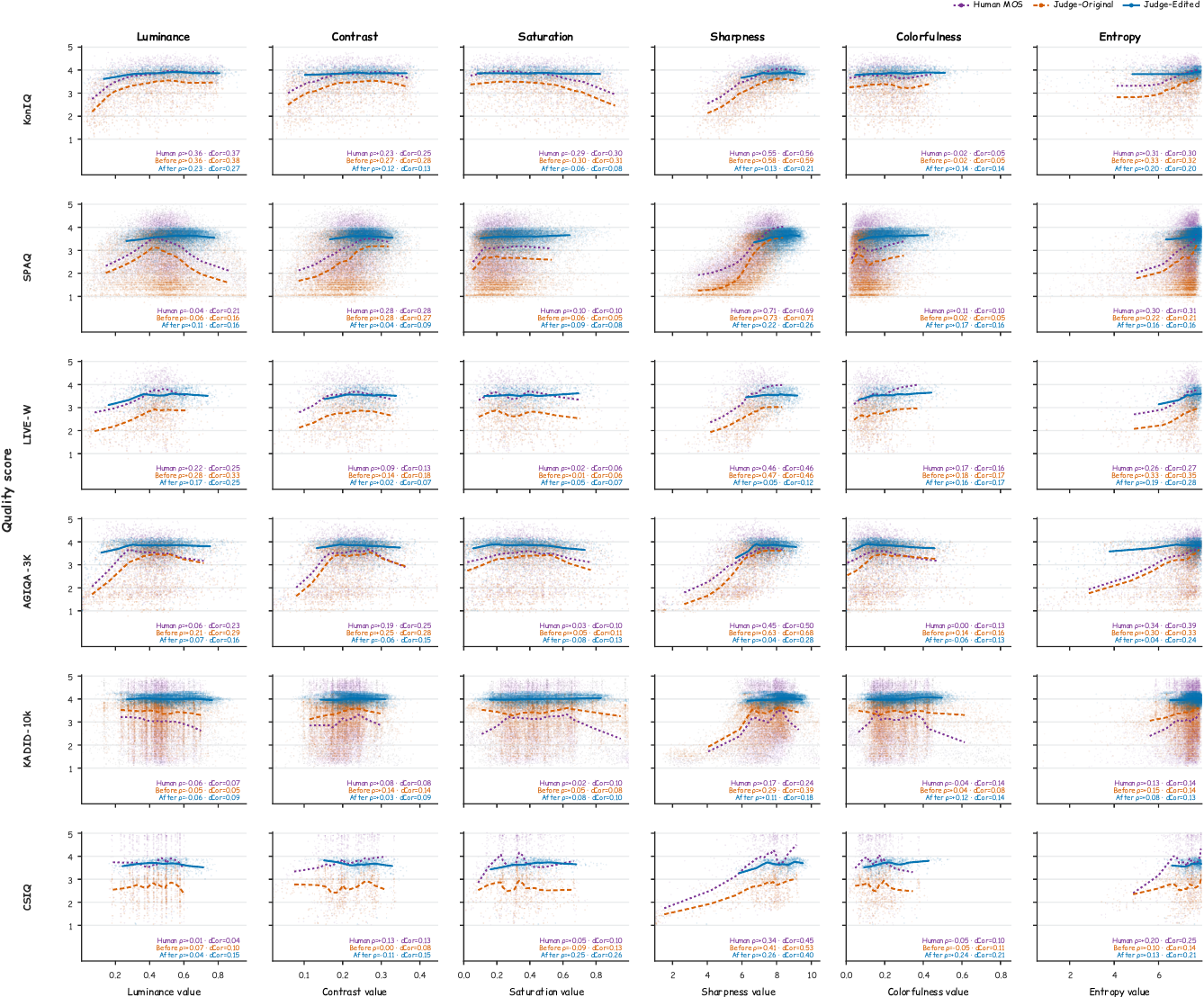}
\caption{\textbf{Quality scores versus low-level image features.} Rows denote
datasets and columns denote luminance, contrast, saturation, sharpness,
colorfulness, and entropy. Purple and orange show human MOS and Judge scores
for original images, while blue shows Judge scores for edited images. Curves
show smoothed trends; $\rho$ and dCor denote Spearman and distance
correlations~\citep{szekely2007distance}.}
\label{fig:quality-low-level-features}
\end{figure*}

\paragraph{Human-like Judge feedback.}
Figure~\ref{fig:quality-low-level-features} shows that the feature-dependent
curves produced by the original-image Judge broadly follow the corresponding
human-MOS curves across the six datasets. Their similar shapes and directions,
particularly for sharpness and contrast, indicate that the Judge captures
human-like quality preferences over low-level visual attributes. This
agreement supports its use as a qualified source of human-like feedback for
the original--edited image pairs.

\paragraph{Cross-dataset feature patterns.}
Among luminance, contrast, saturation, sharpness, colorfulness, and entropy,
sharpness exhibits the clearest cross-dataset trend. Both human MOS and Judge
scores generally increase with sharpness, whereas the other attributes show
weaker, nonlinear, or dataset-dependent patterns. This result suggests that
sharpness is a broadly shared quality cue, while no single low-level attribute
fully explains perceptual quality across all datasets.

\paragraph{Edited-feature distributions.}
After editing, the images retain broad distributions across all six low-level
attributes rather than collapsing to fixed feature values. The framework
therefore does not appear to overfit a single low-level pattern when improving
image quality. Nevertheless, Figure~\ref{fig:quality-low-level-features}
examines each attribute marginally. The joint distribution of low-level
attributes, and its relation to the human visual system and learned perceptual
representations such as CLIP-IQA~\citep{wang2023exploring}, warrants further
investigation.

\subsection{Framework Behavior Preference}

\begin{figure*}[p]
\centering
\includegraphics[
    height=0.90\textheight,
    keepaspectratio,
    pagebox=cropbox
]{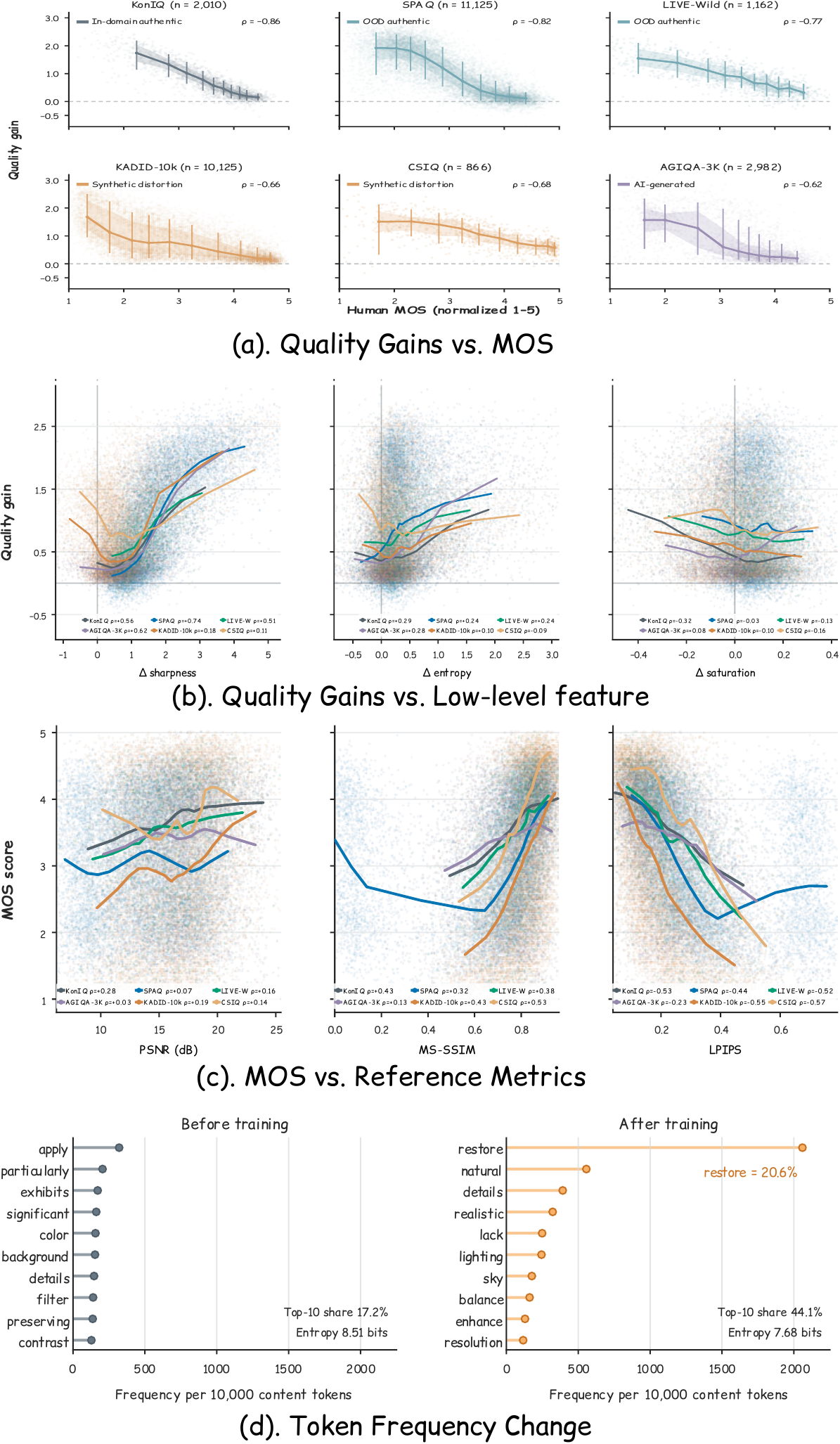}
\caption{\textbf{Behavioral analysis of reasoning-guided editing.}
(a) Judge-rated quality gain versus human MOS across six benchmarks.
(b) Quality gain versus changes in low-level image features.
(c) Human MOS versus reference-based similarity metrics between the original
and edited images.
(d) Frequent reasoning tokens before and after training.}
\label{fig:feature-analysis}
\end{figure*}

\paragraph{Source quality, visual change, and quality gain.}
Figure~\ref{fig:feature-analysis}(a) shows that lower-MOS images generally
receive larger Judge-rated quality gains. Panel (c) further shows that
lower-MOS images tend to have lower similarity between the original and edited
images, indicating larger visual changes. Together with the observation in
Section~\ref{app:quality-fidelity} that larger visual changes are associated
with higher quality gains, these results suggest a plausible behavior:
for lower-quality inputs, the model tends to modify more visual attributes,
which in turn leads to greater quality improvements. This behavior aligns with
the intuition that lower-quality images provide more room for correction and
supports the intended operation of our framework, although the observed
associations do not by themselves establish causality.

\paragraph{Preferences over feature changes.}
Unlike Section~\ref{app:quality-low-level-features}, which examines the
absolute values of low-level attributes, Figure~\ref{fig:feature-analysis}(b)
relates their changes after editing to the resulting quality gain. Changes in
sharpness retain the strongest and most consistent relationship with quality
improvement. For entropy and saturation, the authentic and synthetic datasets
form different response patterns, indicating that the preferred corrections
depend on the degradation domain. SPAQ again exhibits signals that are less
consistent with the other datasets, in agreement with the dataset-specific
behavior observed in Section~\ref{app:quality-fidelity}.

\paragraph{Behavioral adaptation and concentration.}
Figure~\ref{fig:feature-analysis}(d) reflects how the model adjusts its
reasoning vocabulary in response to Judge feedback. After training, terms such
as \emph{natural} and \emph{realistic} become prominent, revealing a learned
preference for perceptually plausible corrections. At the same time,
\emph{restore} alone accounts for 20.6\% of the displayed token frequency, and
the top-10 token share rises to 44.1\%. This concentration suggests an emerging
tendency to overfit a small set of solution patterns, despite the broader
behavioral alignment induced by the framework.

\section{Discussion on the Editor Proxy}
\label{app:editor-proxy}

This work does not disentangle the Editor's instruction-following capability
from semantic preservation, creating two attribution risks. First, quality may
degrade because of Editor limitations rather than an incorrect Actor
instruction. Second, the edited image may become semantically inconsistent
with the input, weakening quality change as a proxy for reasoning faithfulness.
Future work should use independent human or human-like evaluation to assess
instruction following, semantic consistency, and perceptual quality
separately, yielding more reliable feedback and a more faithful framework.

\FloatBarrier
\section{Joint Rating, Reasoning, and Diversity Analysis}
\label{app:joint-rating-reasoning}

\paragraph{Evaluation protocol.}
We jointly examine rating generalization, functional reasoning, and language
diversity. Rating results are six-dataset averages over 28,270 test
images. Functional reasoning is measured by the Judge-score gain $\Delta s$
on the same 5,843 image keys for every Actor, using a fixed FLUX Editor and E5
Judge. This common-case evaluation excludes 23 Mask generations that reached
the length limit. Language diversity is evaluated on 23,599 common successful
inputs. Evidence and solution uniqueness are exact-string statistics, with the
top-1 share measuring the concentration of the most frequent output. Results
should therefore be compared within each metric and its stated sample scope.

\begin{table*}[!t]
\caption{\textbf{Dataset-level rating performance of the credit-routing
variants.} Dataset and Average entries report PLCC/SRCC, where Average is the
unweighted average over the six datasets. The three E1 variants share the same
source checkpoint; the two E5 runs are trained from the native parent for five
epochs. The first two E1 variants use no KL. Bold marks the best result before
rounding in each column.}
\label{tab:dataset-rating-credit-variants}
\centering
{\footnotesize
\setlength{\tabcolsep}{2.0pt}
\renewcommand{\arraystretch}{1.00}
\begin{tabular}{@{}lcccccccc@{}}
\toprule
\textbf{Method}
& \textbf{KonIQ} & \textbf{SPAQ} & \textbf{LIVE-W}
& \textbf{AGIQA-3K} & \textbf{KADID-10k} & \textbf{CSIQ}
& \textbf{Average} & \textbf{MAE} $\downarrow$ \\
\midrule
Without Mask (E1)
& $\mathbf{0.952/0.938}$ & $\mathbf{0.901}/0.898$ & $0.897/0.877$
& $\mathbf{0.816/0.747}$ & $0.661/0.668$ & $0.771/0.746$
& $0.833/\mathbf{0.812}$ & $\mathbf{0.492}$ \\
Mask, no KL (E1)
& $0.945/0.931$ & $0.894/0.892$ & $0.892/0.867$
& $0.805/0.719$ & $0.641/0.650$ & $0.758/0.727$
& $0.822/0.798$ & $0.706$ \\
Mask + KL (E1)
& $0.951/0.938$ & $0.896/0.895$ & $\mathbf{0.903/0.880}$
& $0.801/0.731$ & $0.653/0.664$ & $0.784/0.758$
& $0.831/0.811$ & $0.676$ \\
\midrule
Without Mask (E5)
& $0.932/0.914$ & $0.896/0.897$ & $0.878/0.857$
& $0.809/0.736$ & $\mathbf{0.676/0.684}$ & $0.815/\mathbf{0.786}$
& $0.834/0.812$ & $1.054$ \\
Mask (E5)
& $0.937/0.917$ & $0.900/\mathbf{0.899}$ & $0.893/0.863$
& $0.809/0.739$ & $0.667/0.669$ & $\mathbf{0.824}/0.785$
& $\mathbf{0.838}/0.812$ & $0.543$ \\
\bottomrule
\end{tabular}
}
\end{table*}

\paragraph{Dataset-level rating performance.}
Table~\ref{tab:dataset-rating-credit-variants} provides the complete
cross-dataset results. In the controlled E1 comparison, Without Mask
performs best on KonIQ, SPAQ, and AGIQA-3K and has the lowest average MAE. Mask,
no KL has lower average correlation. Adding KL to Mask recovers most of this gap
and gives the strongest LIVE-W result. At E5, Without Mask is stronger on
KADID-10k and slightly higher in CSIQ SRCC, whereas Mask has the best average
PLCC and is stronger on most authentic and AI-generated benchmarks. Despite their
similar average correlation, Without Mask has a substantially higher MAE
($1.054$ versus $0.543$), exposing a calibration failure that PLCC and SRCC
alone do not capture.

\begin{table*}[!t]
\caption{\textbf{Joint rating, functional reasoning, and diversity analysis.}
$\mathrm{P}/\mathrm{S}$ is the six-dataset average PLCC/SRCC. $\Delta s$ and
Pos. are measured on 5,843 common images. Diversity columns report percentages;
lower top-1 means less concentration. Unmarked diversity values use 23,599
matched inputs. $^\dagger$ marks full E5 audits (28,270 Without Mask; 28,044
Mask).}
\label{tab:joint-rating-reasoning-diversity}
\centering
{\footnotesize
\setlength{\tabcolsep}{2.0pt}
\renewcommand{\arraystretch}{1.00}
\begin{tabular}{@{}lccccccc@{}}
\toprule
& \multicolumn{1}{c}{Rating} & \multicolumn{2}{c}{Functional reasoning}
& \multicolumn{4}{c}{Language diversity} \\
\cmidrule(lr){2-2} \cmidrule(lr){3-4} \cmidrule(lr){5-8}
\textbf{Method}
& \shortstack{$\mathrm{P}/\mathrm{S}$\\$\uparrow$}
& \shortstack{Mean $\Delta s$\\$\uparrow$}
& \shortstack{Pos. (\%)\\$\uparrow$}
& \shortstack{Evidence unique\\(\%) $\uparrow$}
& \shortstack{Evidence top-1\\(\%) $\downarrow$}
& \shortstack{Solution unique\\(\%) $\uparrow$}
& \shortstack{Solution top-1\\(\%) $\downarrow$} \\
\midrule
Baseline
& $0.652/0.645$ & $-0.136$ & $40.664$ & -- & -- & -- & -- \\
Rating-only
& $0.836/\mathbf{0.815}$ & $-0.260$ & $30.686$ & -- & -- & -- & -- \\
Without Mask (E1)
& $0.833/0.812$ & $+0.837$ & $96.954$
& 0.013 & 99.992 & 50.621 & 0.822 \\
Mask, no KL (E1)
& $0.822/0.798$ & $+0.917$ & $97.707$
& 1.233 & 21.196 & \textbf{94.508} & \textbf{0.085} \\
Mask + KL (E1)
& $0.831/0.811$ & $+0.788$ & $95.054$
& 24.018 & 13.191 & 75.228 & 0.191 \\
\midrule
Without Mask (E5)
& $0.834/0.812$ & $\mathbf{+1.431}$ & $\mathbf{99.997}$
& 74.719$^\dagger$ & 0.556$^\dagger$
& 0.004$^\dagger$ & 100.000$^\dagger$ \\
Mask (E5)
& $\mathbf{0.838}/0.812$ & $+1.071$ & $98.813$
& \textbf{83.131} & \textbf{0.407}
& 83.644$^\dagger$ & 0.182$^\dagger$ \\
\bottomrule
\end{tabular}
}
\end{table*}

\paragraph{Rating and functional reasoning.}
Table~\ref{tab:joint-rating-reasoning-diversity} extends the main ablation in
Table~\ref{tab:credit-ablation-rating} with output-level diagnostics. The
rating-only Actor attains strong rating alignment but produces edits that
reduce quality on average. Among the controlled E1 variants, Without Mask has
the strongest rating correlation, whereas Mask gives the largest quality gain.
Mask + KL recovers most of the rating gap while retaining a positive mean gain.
These results separate rating accuracy from the functional usefulness of
reasoning.

\paragraph{Evidence and solution diversity.}
Without Mask produces only three distinct evidence strings, with one string
accounting for 99.992\% of the matched outputs. Its evidence has therefore
collapsed even though its solutions remain more varied. Mask increases the
number of unique evidence strings to 291 and yields the highest solution
uniqueness. Adding KL further raises evidence uniqueness to 5,668 and lowers
its top-1 share to 13.191\%, at the cost of some solution diversity and edit
gain. Evaluating evidence and solution separately is thus necessary: rating
correlation, full-output validity, or solution diversity alone can conceal a
collapsed reasoning field. Exact uniqueness remains a lexical proxy and does
not by itself establish semantic diversity or faithfulness. At E5, Without
Mask retains diverse evidence (74.719\% unique; 0.556\% top-1), while its
solution uniqueness falls to 0.004\% and its solution top-1 share reaches
100\%. This field-wise mismatch shows why evidence and solution must be audited
separately.

\paragraph{Final E5 balance.}
The E5 rows in Table~\ref{tab:joint-rating-reasoning-diversity} reinforce this
multi-objective interpretation. Without Mask E5 reaches a raw gain of $+1.431$
but maps all 28,270 inputs to one normalized solution in the same template
family despite retaining diverse evidence. Mask E5 obtains an average
PLCC/SRCC of $0.838/0.812$, a mean quality gain of $+1.071$, and 19,618 unique
evidence strings on the matched audit. It also produces 23,457 normalized
solutions among 28,044 successful full-audit cases, with no detected
house-template outputs. The higher Without Mask reward therefore reflects a
universal edit rather than image-conditioned solution reasoning.
Because the final runs change both credit routing and KL topology and use one
seed per setting, this comparison is diagnostic rather than a pure causal
estimate of the credit mask. Section~\ref{app:credit-mask-audit} provides the
corresponding training and validation dynamics.

\section{Detailed Credit-Mask Audit}
\label{app:credit-mask-audit}

\subsection{Audit Scope and Protocol}
We audit two complete five-epoch runs with the same Actor, data, prompt, and
GRPO settings. \textbf{Mask} routes each reward to its supervised output with
output-specific KL regularization; \textbf{Without Mask} broadcasts rewards
over the complete output with one global KL term. Both use a KL coefficient of
$0.02$. The diagnostic reasoning reward bounds the raw Judge-score change:
\begin{equation}
    R_{i,k}^{\mathrm{diag}}
    =\operatorname{sgn}(\Delta s_{i,k})
    \left(1-\mathrm{e}^{-\Delta s_{i,k}^{2}/2}\right).
    \label{eq:diagnostic-reasoning-reward}
\end{equation}
The Editor consumes only the solution without an explicit semantic guardrail.
Both runs use six samples per image, 160-token completions, and frozen vision
encoder and aligner. Each contains 1,455 steps with 144 trajectories per step
(209,520 per run; 419,040 total). No step or record is missing, duplicated,
invalid, or non-finite. At Without Mask steps 711 and 713, all samples are
Actor-ineligible, so absent Judge changes reflect eligibility rather than
numerical or service failure. Because credit and KL routing change jointly and
each setting has one seed, this audit is descriptive rather than causal.

\subsection{Training Dynamics}
\begin{figure*}[!t]
\centering
\includegraphics[width=0.85\textwidth]{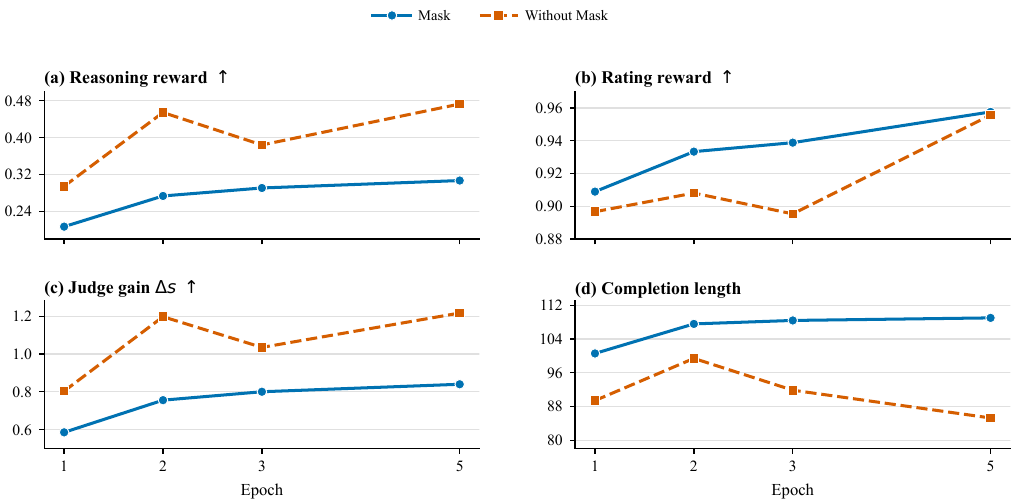}
\caption{\textbf{Training dynamics at the retained checkpoints.} Each point is
the exact four-rank mean for E1, E2, E3, or E5. Without Mask produces larger
reasoning reward and Judge gain, whereas rating reward converges similarly.
Its concurrent reduction in completion length is consistent with convergence
toward a high-reward solution template.}
\label{fig:credit-mask-training}
\end{figure*}

Figure~\ref{fig:credit-mask-training} reports exact four-rank means at the
retained checkpoints. Both configurations improve rating reward. Without Mask
produces larger reasoning reward and Judge gain, but also shorter outputs and
solution collapse. Mask changes more gradually: its reasoning reward rises
from 0.207 to 0.307 and Judge gain from 0.586 to 0.840. Rating rewards converge
by E5. KL losses are not directly comparable because their token scopes differ.
For Without Mask, mean reasoning reward and Judge gain increase from 0.463 and
1.189 over steps 1256--1355 to 0.478 and 1.232 over steps 1356--1455. Yet their
final-window slopes are $-0.0108$ and $-0.0246$; the rating-reward slope is
$-0.0006$. The endpoint therefore masks a plateau and slight decline.

\subsection{Checkpoint Validation}
\begin{figure*}[!t]
\centering
\includegraphics[width=0.85\textwidth]{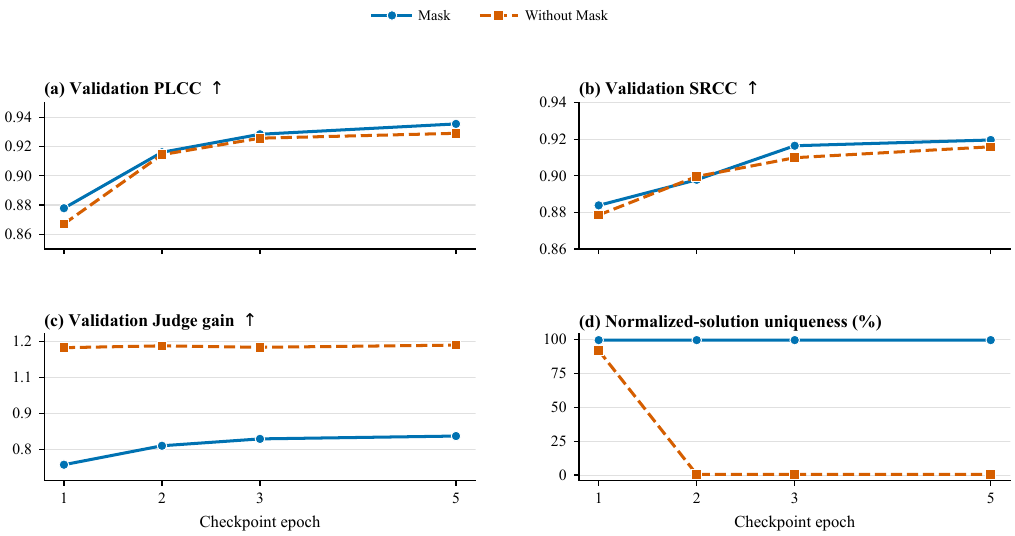}
\caption{\textbf{Checkpoint validation dynamics.} Panels (a)--(c) report
PLCC, SRCC, and Judge gain on the complete 200-image validation set. Panel (d)
reports normalized-solution uniqueness. Without Mask maintains rating alignment
and high Judge gain while its solutions collapse.}
\label{fig:credit-mask-validation}
\end{figure*}

Figure~\ref{fig:credit-mask-validation} evaluates the retained checkpoints on
the same 200 images. PLCC and SRCC improve for both runs. Without Mask keeps a
Judge gain near 1.18, while normalized-solution uniqueness falls from 91.88\%
at E1 to about 0.5\% at E2. The audit further shows that the semantic
dwelling-template rate reaches 100\% at E1. Semantic collapse therefore
precedes near-exact lexical collapse.

\subsection{Cross-Dataset Outcomes}
At E5, Without Mask obtains a larger pooled quality gain than Mask
($1.431$ versus $1.071$), although their six-dataset PLCC/SRCC averages remain
close ($0.834/0.812$ versus $0.838/0.812$). Larger gain alone therefore does not
establish healthier reasoning. All 28,270 Without Mask outputs share one
normalized solution; Mask retains 23,457 among 28,044 successful outputs.

\subsection{Collapse Milestones}
Table~\ref{tab:completion-audit-milestones} distinguishes semantic,
phrase-level, and near-exact collapse. Under Without Mask, the semantic
dwelling family exceeds 90\% at step 265, whereas one normalized sentence does
not exceed 90\% until step 551. Duplicate matching would detect the failure 286
steps late. Mask avoids dwelling collapse, although its super-resolution and
smoothing phrase backbone remains above 90\% from step 407 onward.

\begin{center}
\begin{minipage}{\columnwidth}
\centering
{\scriptsize
\setlength{\tabcolsep}{2.5pt}
\begin{tabular}{@{}p{1.55in}rrl@{}}
\toprule
\textbf{Routing and detector}
& $\boldsymbol{\geq 50\%}$ & $\boldsymbol{\geq 90\%}$
& \textbf{Sustained} \\
\midrule
Without Mask: semantic dwelling & 236 & 265 & 948--E5 \\
Without Mask: strict target phrase & 356 & 475 & -- \\
Without Mask: modal normalized solution & 520 & 551 & 973--E5 \\
Without Mask: canonical sentence & 1,133 & 1,146 & -- \\
Mask: super-resolution + smoothing & 214 & 260 & 407--E5 \\
\bottomrule
\end{tabular}
}
\captionof{table}{\textbf{Collapse milestones in global optimizer steps.} Semantic
detectors identify shared concepts before normalized near-duplicate matching
identifies one sentence.}
\label{tab:completion-audit-milestones}
\end{minipage}
\end{center}

\subsection{Output-Length Evolution}
Panel (d) of Figure~\ref{fig:credit-mask-training} shows that the mean training
completion decreases from 89.39 to 85.26 tokens under Without Mask but rises
from 100.57 to 109.00 under Mask. On validation outputs, the corresponding
solution lengths change from 32.56 to 18.00 and from 37.85 to 50.94. The former
contraction is consistent with convergence to one reusable instruction.

\subsection{Rating and Diversity Diagnostics}
Without Mask retains a KonIQ PLCC/SRCC of $0.932/0.914$ despite solution
collapse. Ratings still vary: 2,009 of 2,010 complete JSON outputs are unique,
although all 2,010 solutions are identical after normalization. Rating
correlation and full-output uniqueness therefore miss solution collapse.
Evaluation must inspect the supervised output alongside image--solution
relevance and semantic preservation.

\subsection{Representative Validation Outputs}
The Without Mask E5 Actor produces different evidence and ratings for the
three examples, but all solutions normalize to one dwelling instruction.
Sample 1 still receives 0.865 reward because the fixed intervention raises the
Judge score from 2.37 to 4.37. Thus, the reward measures edited-image quality
without establishing source-image faithfulness.

\begin{center}
\begin{minipage}{\columnwidth}
\centering
{\small
\setlength{\tabcolsep}{5pt}
\begin{tabular}{@{}lrrrrr@{}}
\toprule
\textbf{Sample} & \textbf{Rating} & $\boldsymbol{J_0}$
& $\boldsymbol{J_1}$ & $\boldsymbol{\Delta s}$ & $\boldsymbol{R}$ \\
\midrule
1 & 1.68 & 2.37 & 4.37 & 2.00 & 0.865 \\
2 & 3.32 & 3.84 & 4.32 & 0.48 & 0.109 \\
3 & 2.65 & 3.02 & 4.36 & 1.34 & 0.593 \\
\bottomrule
\end{tabular}
}
\captionof{table}{Representative Without Mask validation outputs. Different evidence
and ratings lead to the same normalized solution.}
\label{tab:completion-audit-examples}
\end{minipage}
\end{center}

\subsection{Causal Scope}
The comparison changes credit and KL routing jointly with one seed per setting.
It establishes a reproducible failure mode for Without Mask/global KL, but not
the responsible change. Causal attribution requires a $2{\times}2$ grid of
Without Mask versus Mask and global versus component KL, with multiple seeds.

\FloatBarrier

\end{document}